\documentclass[pdflatex,sn-mathphys-num]{sn-jnl}
\usepackage{graphicx}%
\usepackage{multirow}%
\usepackage{amsmath,amssymb,amsfonts}%
\usepackage{amsthm}%
\usepackage{mathrsfs}%
\usepackage[title]{appendix}%
\usepackage{xcolor}%
\usepackage{textcomp}%
\usepackage{manyfoot}%
\usepackage{booktabs}%
\usepackage{algorithm}%
\usepackage{algorithmicx}%
\usepackage{algpseudocode}%
\usepackage{listings}%

\theoremstyle{thmstyleone}%
\theoremstyle{thmstyletwo}%

\theoremstyle{thmstylethree}%
\begin{document}

\title[Generalizing Deep Surrogate Solvers for Broadband Electromagnetic Field Prediction at Unseen Wavelengths]{Generalizing Deep Surrogate Solvers for Broadband Electromagnetic Field Prediction at Unseen Wavelengths}

\author[1]{\fnm{Joonhyuk} \sur{Seo}}\email{yhy258@hanyang.ac.kr}
\equalcont{These authors contributed equally to this work.}

\author[1]{\fnm{Chanik} \sur{Kang}}\email{chanik@hanyang.ac.kr}
\equalcont{These authors contributed equally to this work.}

\author[2]{\fnm{Dongjin} \sur{Seo}}\email{}

\author*[1]{\fnm{Haejun} \sur{Chung}}\email{haejun@hanyang.ac.kr}

\affil*[1]{\orgdiv{Department of Artificial Intelligence}, \orgname{Hanyang University}, \orgaddress{\street{222 Wangsimni-ro}, \postcode{04763}, \state{Seoul}, \country{Republic of Korea}}}

\affil[2]{\orgdiv{Department of Applied Physics}, \orgname{Yale University}, \orgaddress{\street{55 College St}, \city{Connecticut}, \postcode{06520}, \state{New Haven}, \country{USA}}}

\abstract{
Recently, electromagnetic surrogate solvers, trained on solutions of Maxwell’s equations under specific simulation conditions, enabled fast inference of computationally expensive simulations. However, conventional electromagnetic surrogate solvers often consider only a narrow range of spectrum and fail when encountering even slight variations in simulation conditions. To address this limitation, we define spectral consistency as the property by which the spatial frequency structure of wavelength-dependent condition embeddings matches that of the target electromagnetic field patterns. In addition, we propose two complementary components: a refined wave prior, which is the condition embedding that satisfies spectral consistency, and Wave-Informed element-wise Multiplicative Encoding (WIME), which integrates these embeddings throughout the model while preserving spectral consistency. This framework enables accurate field prediction across the broadband spectrum, including untrained intermediate wavelengths. Our approach reduces the normalized mean squared error at untrained wavelengths by up to 71\% compared to the state-of-the-art electromagnetic surrogate solver and achieves a speedup of over 42 times relative to conventional numerical simulations.
}

\keywords{Surrogate Solver, Neural Operator, Electromagnetics, Nanophotonics}

\maketitle

\section{Introduction}\label{sec1}

Recent advances at the intersection of deep learning (DL) and photonics have led to significant breakthroughs across various domains, including the inverse design of photonic structures~\cite{wen2019progressive, chen2022high, seo2025physics, seo2024high, kang2025adjoint}, high-resolution imaging and computational microscopy~\cite{seo2023deep, tseng2021neural, deb2022fouriernets}, optical signal processing and communication systems~\cite{tan2023photonic, fan2020advancing, huang2021all}, the realization of optical neural networks~\cite{lin2018all, xu202111, fu2024optical}, and a digital twin of the physical system~\cite{d2022physics, wright2022deep, momeni2023backpropagation}. Specifically, DL-based approaches have demonstrated remarkable capabilities in rapidly predicting electromagnetic phenomena traditionally analyzed through numerical methods~\cite{jiang2021deep, chen2022high, so2020deep, gu2022neurolight}. By training neural networks on datasets derived from simulations or experiments, DL approaches achieve orders-of-magnitude acceleration in predicting electromagnetic fields, device performance, and structural optimizations, all while maintaining accuracy comparable to traditional numerical solvers~\cite{jiang2021deep, chen2022high, so2020deep, gu2022neurolight}. 

One important application of DL in photonics is the development of surrogate solvers that approximate electromagnetic simulations. These surrogate solvers predominantly utilize deep neural networks due to their rapid inference and notable learning capability. Traditional data-driven neural network-based surrogate solvers~\cite{malkiel2018plasmonic, peurifoy2018nanophotonic, sajedian2019finding, so2020deep, jiang2021deep} have attempted to learn Maxwell’s equations by training on simulated electromagnetic field data. However, these approaches frequently encounter limitations in accurately capturing the underlying physical characteristics of Maxwell’s equations~\cite{chen2022high}. 

To address these shortcomings, physics-informed neural networks (PINNs)~\cite{lu2021physics, zhelyeznyakov2023large, piao2024domain} and methods integrating physics-based loss functions~\cite{chen2022high, lim2022maxwellnet} have been extensively explored. These methods incorporate Maxwell's equations directly into the loss functions used during neural network training, thereby enabling the model to better capture the electromagnetic characteristics imposed by Maxwell’s equations. Although these physics-informed approaches significantly improve prediction speeds compared to conventional numerical solvers, they still face important challenges. Specifically, they require distinct datasets for each simulation parameter, such as wavelength and material permittivity. Therefore, they typically exhibit reduced field prediction accuracy for untrained simulation parameters.

Recently, surrogate solvers utilizing neural operators have gained attention, as these methods effectively learn mappings between infinite-dimensional function spaces, reflecting the intrinsic characteristics of Partial Differential Equations (PDEs)~\cite{kovachki2023neural, azizzadenesheli2024neural}. Notable progress in operator learning includes transformer-based neural operators~\cite{cao2021choose, hao2023gnot, li2022transformer, li2024scalable}, convolutional neural operators~\cite{raonic2023convolutional}, and Fourier neural operators (FNO)~\cite{azizzadenesheli2024neural, li2020fourier, tran2021factorized, li2023fourier, li2024geometry, gu2022neurolight}. Among these, FNO has demonstrated strong inference performance across various electromagnetic applications~\cite{augenstein2023neural, mao2024towards, gu2022neurolight}, making it particularly promising for accelerating electromagnetic simulations.

Nevertheless, recent surrogate solver studies, including FNO-based methods, exhibit limited prediction accuracy under the untrained simulation conditions, thereby failing to demonstrate the generalization capability of the  models (a related assessment is presented in Sections 2.3 and 2.4). The notable decline in predictive accuracy at intermediate wavelengths between discretely sampled training points raises fundamental concerns about whether these models adequately capture the underlying physics governing electromagnetic phenomena. 

In this work, we present an electromagnetic surrogate solver for achieving highly accurate full-wave Maxwell solutions across a continuous and broadband spectrum, even when training data is available only at sparsely sampled, discrete wavelengths. To this end, we introduce two key concepts: (i) the concept of \textit{Spectral Consistency}, which captures the intrinsic relationship between wavelength-dependent variations and spatial frequency governed by Maxwell’s equations, and (ii) a conditional embedding strategy alongside an effective conditioning method. Specifically, we propose \textit{Refined Wave Prior} and \textit{Wave-Informed Element-wise Multiplicative Encoding (WIME)}, designed to explicitly preserve spectral consistency. By integrating these ideas, our model achieves robust and accurate predictions across a wide range of untrained wavelengths, addressing the limitations of existing surrogate solvers and significantly advancing the generalization capability of deep-learning-based electromagnetic simulators.

\section{Results}\label{sec2}

\begin{figure*}[!htp]
\centering
\includegraphics[width=1.0\columnwidth]{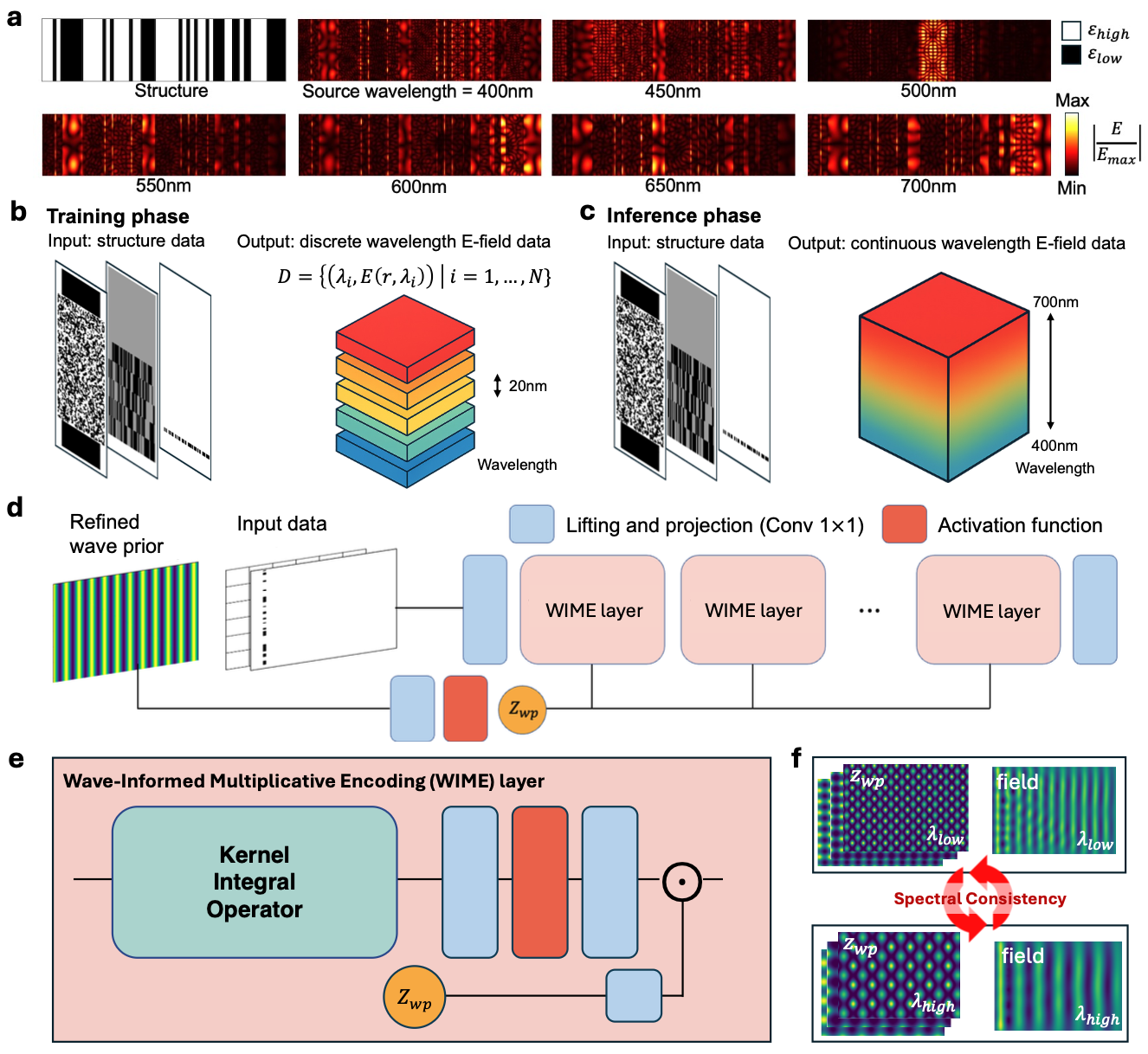} 
\caption{\textbf{Overview of our study} \textbf{a} Schematic of 2D subwavelength gratings, and their electric field profile under normal incidence plane wave. \textbf{b} Training phase: Both permittivity profile and electric field profile over the simulation domain are used in neural network training. The discrete wavelength electric field data is uniformly spaced by 20 nm over the visible wavelength range. \textbf{c} Inference phase: for given permittivity profile and wavelength, the trained model predicts accurate field distributions even for untrained wavelengths. \textbf{d} Network architecture: a refined wave prior (condition embedding) and the permittivity profile (input data) are lifted via 1$\times1$ convolutions (blue), then processed through repeated Wave-Informed Multiplicative Encoding (WIME) layers (pink), with latent refined wave-prior embeddings $Z_{wp}$ injected at every layer. \textbf{e} Structure of a single WIME layer: inputs are first transformed by a kernel integral operator (green), then are processed by 1×1 convolutions (blue) and a nonlinear activation (red); the projected wave-prior embedding is element-wisely multiplied ($\odot$) with the feature map. \textbf{f} Spectral consistency: for different wavelengths, the latent refined wave prior (left) and the corresponding predicted electric field patterns (right) show similar spatial frequency characteristics.}
\label{fig1}
\end{figure*}

\subsection{Nonlinear variation of field patterns over wavelength shift}

In subwavelength photonic structures, various electromagnetic phenomena such as resonance~\cite{zhang2020quasinormal}, scattering~\cite{ruan2010superscattering}, and diffraction~\cite{goodman2005introduction} may complicate the accurate prediction or generalization of electromagnetic fields at untrained wavelengths using deep learning models. Furthermore, these electromagnetic phenomena become more pronounced in regions with high permittivity contrast~\cite{meade2008photonic}. Such contrast with randomly arranged structures can intensify field localization (e.g., through the formation of evanescent waves at dielectric interfaces) and promotes multiple scattering events~\cite{herzig2016interplay}. In addition, the geometrically unconstrained designs of freeform structures can further lead to intricate field patterns, which may introduce additional electromagnetic phenomena like effective medium effects~\cite{lalanne1996effective}. The intricate interplay results in field profiles that exhibit strong nonlinear variations across wavelengths.

Figure~\ref{fig1}a illustrates the nonlinear variation of electromagnetic field patterns across wavelengths within the design region of the metalens simulation. Details of the simulation configurations are provided in Section 7 of Supplementary Information. The results underscore the challenge of capturing broadband nonlinear variations using conventional surrogate models that rely on discrete wavelength training data. In the following sections, we present our solutions to this challenge, focusing on robust field prediction across the continuous broadband spectrum, including intermediate and untrained wavelengths.

\subsection{Spectral consistency and Wave-Informed element-wise Multiplicative Encoding}

As we introduced in the previous section, complex optical phenomena, such as scattering and resonance, introduce nonlinear wavelength-dependent variations in electric field profiles. Thus, they may hinder accurate prediction at untrained wavelengths. To address this challenge, it is crucial to ensure that the model can incorporate physical characteristics corresponding to arbitrary wavelengths, including untrained ones. This can be achieved by constructing condition embeddings that accurately represent these wavelength-dependent physical features and by designing conditioning methods that effectively integrate them into the model.

Specifically, we note that shorter wavelengths tend to produce field patterns with higher spatial frequencies, whereas longer wavelengths are associated with sparser and lower spatial frequency field patterns. This inverse relationship is an intrinsic feature of Maxwell’s equation solutions and can be interpreted from the relation $\nu = 1/\lambda$, where $\lambda$ and $\nu$ respectively denote the wavelength and spatial frequency~\cite{guenther2018modern}. We define \textit{Spectral Consistency} as the property by which wavelength-dependent condition embeddings, reflecting the inverse relationship, exhibit spatial frequency structures that aligns with those of the desired electromagnetic field patterns.
In this context, condition information satisfying spectral consistency may capture the expected field complexity and spatial frequency distribution for any given wavelength, including those not observed during training. Thus, ensuring spectral consistency can significantly aid surrogate models in generalizing across the broadband spectrum.

We propose conditioning embeddings that accurately capture the wavelength-dependent characteristics and satisfy spectral consistency. First, we employ and refine a wave prior~\cite{gu2022neurolight}. The wave prior $\mathcal{W} = [\mathcal{W}_x, \mathcal{W}_y]$ is an artificial wave pattern derived from the solution of the wave equation where $\mathcal{W}_x$ and $\mathcal{W}_y$ are expressed as $\mathcal{W}_x = e^{j \frac{2\pi \sqrt{\boldsymbol{\epsilon}_r}}{\lambda} x \mathbf{1}^T\Delta l_x}$ and $\mathcal{W}_y = e^{j \frac{2\pi \sqrt{\boldsymbol{\epsilon}_r}}{\lambda} \mathbf{1} y^T \Delta l_y}$. Here, $\boldsymbol{\epsilon}_r \in \mathbb{C}^{H \times W}$ represents the relative permittivity of the structures at each coordinate, $\lambda$ is the wavelength, $\Delta l_x$ and $\Delta l_y$ denote the simulation step sizes. The variables $x$ and $y$ correspond to the spatial coordinates. Since the wave prior is derived from the solution of the wave equation in homogeneous media, it resembles the wave patterns of the fields generated by the simulator, which result from light propagation according to wavelength.

However, note that the wave prior considers only the permittivity values of the structures and does not account for complex electromagnetic phenomena, as it is defined by just substituting coordinates without modeling electromagnetic interactions. Consequently, it exhibits sharp and physically implausible features in regions where the permittivity changes (detailed in Section 4.2 in the Supplementary Information and Fig.~S3).

Hence, we propose a refined wave prior that excludes the $\epsilon_r$ term, defined as $\mathcal W_x = e^{j{2\pi\over \lambda}x \textbf{1}^T\Delta l_x}$ and $\mathcal W_z = e^{j{2\pi\over \lambda}\textbf{1}z^T\Delta l_z}$. The refined condition embedding provide a more physically plausible representation of light behavior in free space (Table S2 and Fig. S3).

In addition, for effective generalization to untrained wavelengths across the broadband spectrum, it is important to employ a conditioning mechanism that integrates condition embeddings while preserving spectral consistency. To this end, we propose Wave-Informed element-wise Multiplicative Encoding (WIME), a method specifically designed to incorporate wavelength-dependent condition information into the model and maintain spectral consistency.

Under the assumption that the given condition embedding satisfies spectral consistency, preserving its spatial structure throughout the network becomes critical to ensuring that the hidden representations of these embeddings maintain spectral consistency as well. To achieve this, we employ only channel-wise operations through $1 \times 1$ convolutional layers, thereby retaining the spatial structure during transformations.

Furthermore, the transformed embeddings are incorporated into each layer of the model via element-wise multiplication with the intermediate feature maps after the kernel integral operator and additional convolutional layers. Details of the kernel integral operator are explained in Section 2.1 of Supplementary Information.
This multiplicative integration enables the model to inject condition information directly and without compression, preserving the fidelity of the wavelength-specific wave characteristics. Consequently, WIME effectively regularizes the physically plausible spatial–spectral structure of the field patterns across varying wavelengths. Further details on the conditioning method are provided in Section 2.2 of the Supplementary Information.

\subsection{Quantitative evaluation on broadband field prediction across various simulations}

\begin{figure*}[htp]
\centering
\includegraphics[width=1.0\columnwidth]{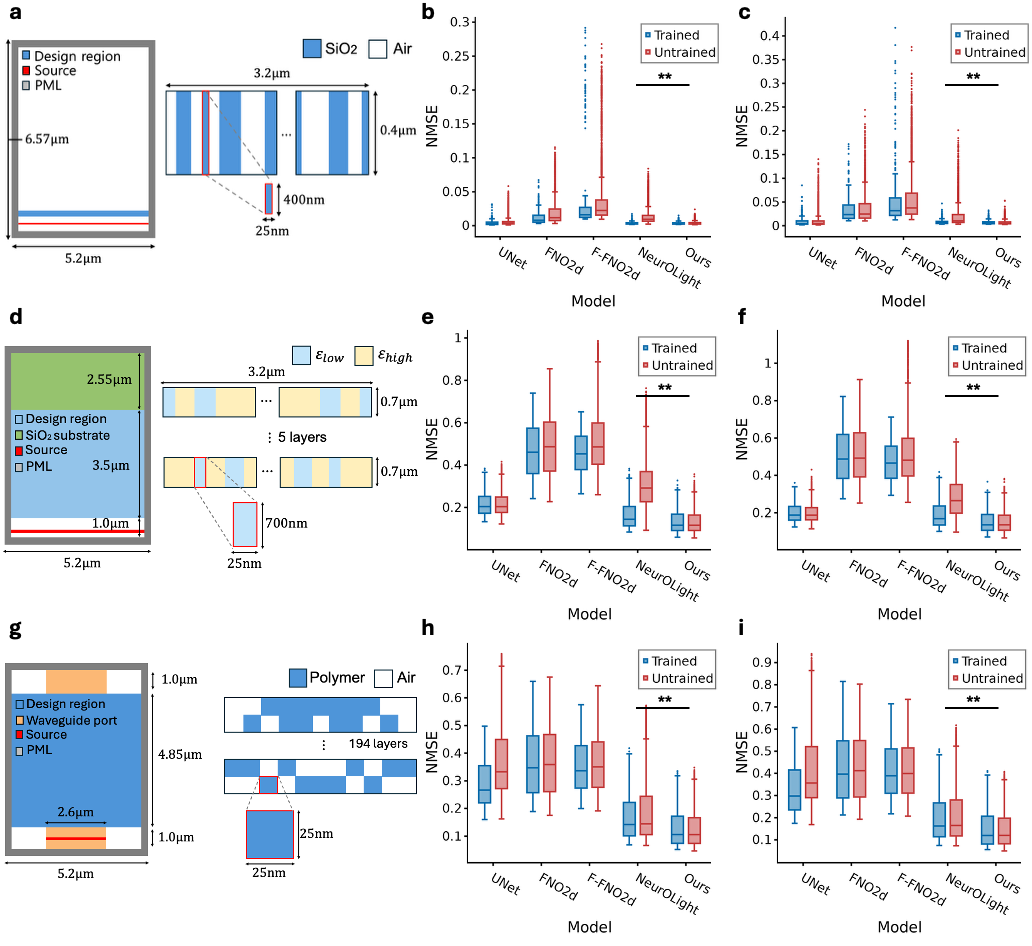} 
\caption{\textbf{Configurations of three optical simulations and quantitative performance comparison across different simulation settings.} \textbf{a, d, g.} Schematics of the three photonic structures: \textbf{a.} single layer metalens; \textbf{d.} multilayer spectrum splitter; and \textbf{g.} freeform straight waveguide. \textbf{b–c.} NMSE distribution for each model under the metalens simulation, evaluated over the entire domain (b) and the design region (c). \textbf{e–f.} Performance under the spectrum splitter configuration, measured across the entire region (e) and the design region (f). \textbf{h–i.} Results for the straight waveguide simulation over the entire domain (h) and design region (i). Asterisks (\textbf{**}) indicate statistically significant differences (P \(<\) 0.01) between NeurOLight and our model based on the Wilcoxon signed-rank test, as the distributions were non-Gaussian. In addition, ``Trained'' and ``Untrained'' refer to the prediction results for trained and untrained wavelengths, respectively.}
\label{fig2}
\end{figure*}

In the following, we compare the field prediction performance for the continuous and broadband spectrum, including untrained wavelengths, in various simulation environments. We select three representative optical structures to investigate the generalization performance of our model under diverse electromagnetic phenomena. 

The first example is a single layer metalens~\cite{park2019all, chung2020high, bayati2020inverse}, which features single layer and thin subwavelength structures, encompassing both near- and far-field regions. The second structure is the spectrum splitter~\cite{zou2022pixel, lee2024inverse, catrysse2022subwavelength}, where its design region consists of 5 layers with higher contrast in permittivity values compared to the metalens case. The third structure is a straight waveguide~\cite{piggott2015inverse, bae2024topology, chen2024freeform} featuring a fully freeform design region composed of two materials that introduce higher permittivity contrast compared to that of the metalens simulation. These geometrically unconstrained designs and the higher contrast in permittivity of the last two environments can induce more complex electromagnetic phenomena in randomly arranged structures, as explained in Section 2.1. Details of the simulation configurations are provided in Section 7 of Supplementary Information.

In addition, we compare the field prediction performance of UNet~\cite{ho2020denoising, gupta2022towards}, FNO2d~\cite{li2020fourier}, and F-FNO2d~\cite{tran2021factorized}, commonly used in surrogate solver study. Moreover, we evaluate our model against NeurOLight~\cite{gu2022neurolight}, the state-of-the-art model for field prediction of a narrow wavelength range in EM simulation. For several baseline models (UNet, FNO2d, F-FNO2d), the conditional information was incorporated by simply concatenating the refined wave prior to the input. NeurOLight similarly utilized the wave prior concatenation, following the methodology proposed in the original study. Section 6 in Supplementary Information provides further architectural details for all models used in the experiments.

Figure~\ref{fig2}b and c quantitatively compare the field prediction accuracy of the proposed model against various baselines under the single layer metalens simulation setting, encompassing both near- and far-field regions. Across both the entire region and the structure region, the proposed model consistently demonstrates the lowest error for both trained and untrained wavelengths, outperforming all baseline models. The performance gap becomes particularly pronounced under untrained wavelength conditions, highlighting the superior generalization capability of our approach. Even when compared to high-performing models such as UNet and NeurOLight, our model exhibits lower variance in prediction errors, especially at untrained wavelengths. These results indicate that the proposed model effectively captures both the relatively smooth wavelength-dependent variations in the far-field and the more intricate spatial variations in the near-field.

A similar comparison is performed under the spectrum splitter simulation, as illustrated in Fig.~\ref{fig2}e and f. Figure~\ref{fig2}e presents the results across the entire domain, while Fig.~\ref{fig2}f focuses on the design region. In this scenario, the use of materials with a higher refractive index contrast than in the metalens case induces significantly complex and stronger light-matter interaction. These interactions intensify the nonlinearity of electromagnetic field responses with respect to wavelength, substantially challenging field prediction and weakening the assumed spectral consistency.
Despite the challenges in the spectrum splitter setting, our model consistently achieves the lowest prediction errors across both trained and untrained wavelengths (Fig.~\ref{fig2}e). This robustness is evident given that even high-performing baselines such as UNet and NeurOLight experience higher error growth in this scenario. Moreover, Figure~\ref{fig2}f indicates that the proposed method preserves both high accuracy and stability even in the design regions. Compared to the metalens setting, these results highlight the superior generalization and stability of our approach when faced with more complex light–matter interactions. In conclusion, this demonstrates that our method can generalize reliably across wavelengths, even in environments where the spectral consistency assumption is significantly weakened.

Figure~\ref{fig2}h and i show the evaluation for the freeform straight waveguide structure. Although this simulation setting includes a relatively smaller permittivity contrast compared to the spectrum splitter simulation, the unconstrained geometric design causes complex electromagnetic phenomena.
In spite of the increased complexity, the proposed model still achieves the lowest average NMSE and error variance across both the entire region (Fig.~\ref{fig2}h) and the design region (Fig.~\ref{fig2}i). In particular, the error increase for untrained wavelengths remains minimal, further validating the model’s generalization ability. In contrast, the other models exhibit both higher average errors and broader error distributions. The results also demonstrate that our method enables robust electromagnetic field prediction even under the weakened spectral consistency assumption by a highly complex electromagnetic interplay from structurally unconstrained design conditions.

\subsection{Continuous-spectrum generalization and spatial error analysis}
\begin{figure*}[hbt!]
\centering
\includegraphics[width=1.0\columnwidth]{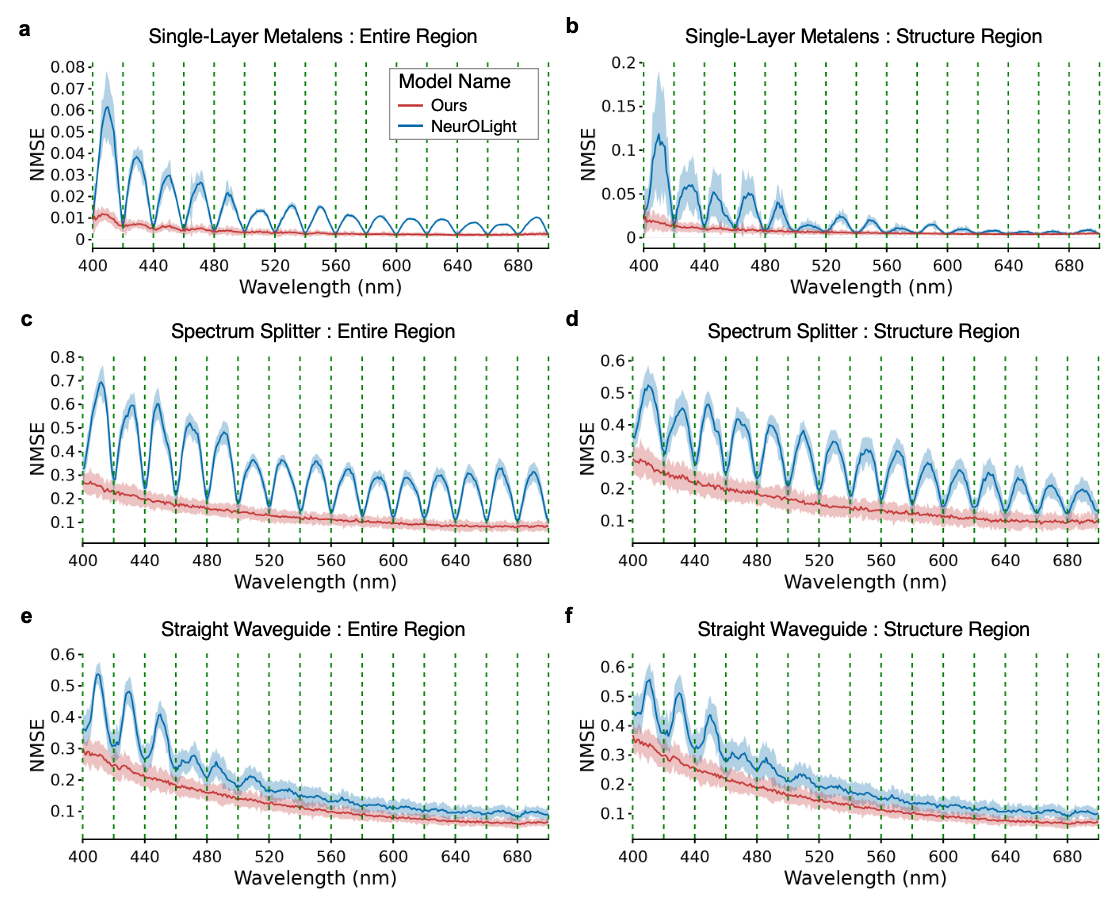} 
\caption{\textbf{Continuous-spectrum generalization and spatial error analysis across various photonic structures.} \textbf{a-f.} Quantitative comparison of our method and NeurOLight across the continuous broadband spectrum, including both trained and untrained wavelengths, under three different simulation settings: \textbf{a-b.} single layer metalens evaluated over (a) the entire domain and (b) the design region; \textbf{c–d.} spectrum splitter evaluated over (c) the entire region and (d) the design region; \textbf{e–f.} straight waveguide evaluated over (e) the entire domain and (f) the design region. The green dashed lines indicate the trained wavelengths. Shaded areas represent ±2 standard deviations for improved visibility, and green dashed lines indicate wavelengths used during training.}
\label{fig3}

\end{figure*}

\begin{figure*}[htbp!]
\centering
\includegraphics[width=1.0\columnwidth]{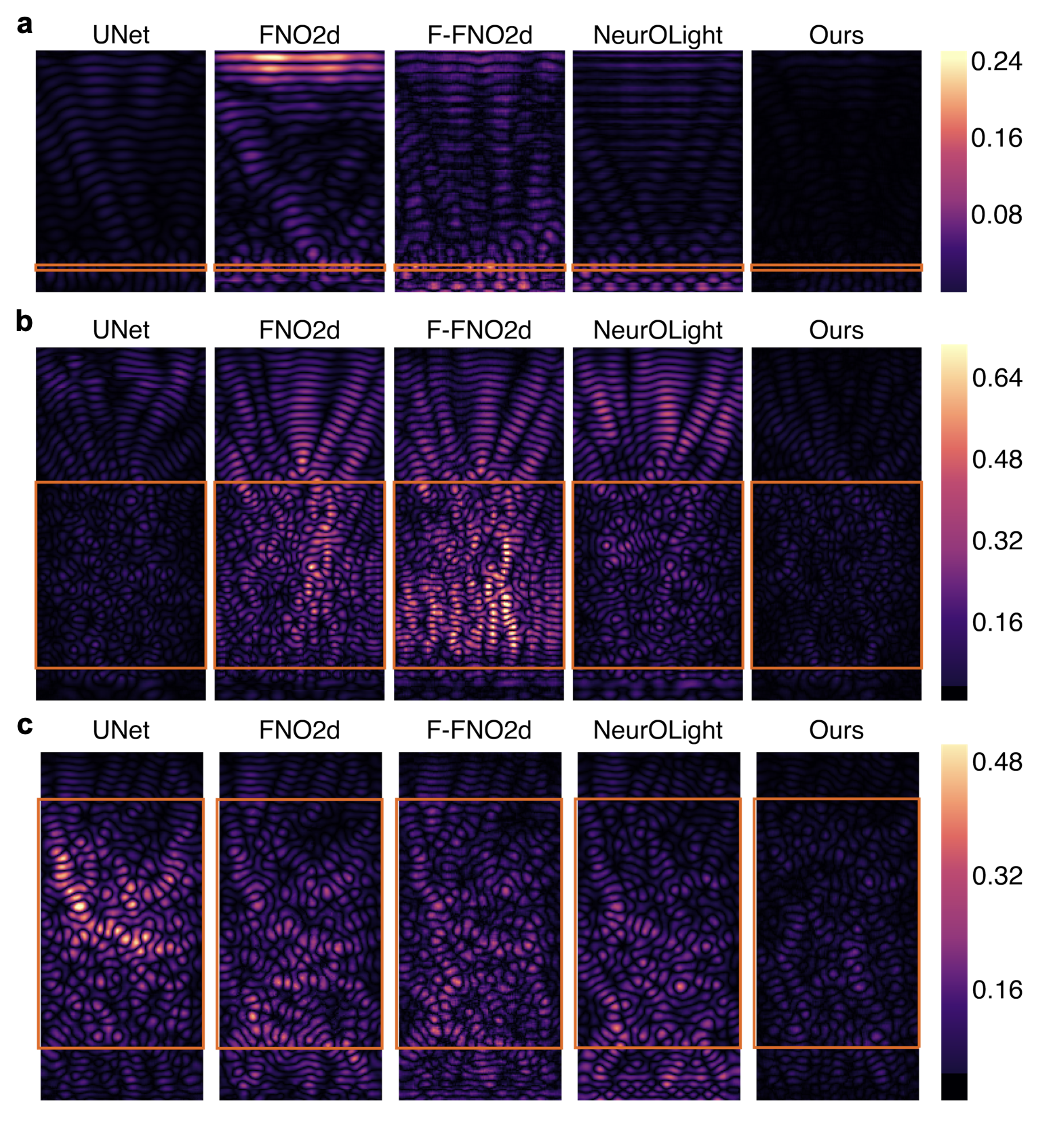} 
\caption{Error maps at 410nm, an unobserved wavelength with the highest prediction error. \textbf{a} the single layer metalens, \textbf{b} the spectrum splitter, and \textbf{c} the straight waveguide. Each column corresponds to a different model. The orange rectangular bounding boxes represent the design region where the strong light-matter interaction emerges.}
\label{fig4}

\end{figure*}

A primary objective of this study is to develop a surrogate solver capable of accurately predicting electric field distributions across a continuous broadband spectrum, including untrained wavelengths.  Accordingly, to rigorously assess the model’s generalization performance, we plot the normalized NMSE across the 400–700nm wavelength range.

Figure~\ref{fig3}a and b visualize the prediction errors over the broadband continuous wavelength range for the single layer metalens setting illustrated in Fig.~\ref{fig2}a. The green dashed lines in the figure indicate trained wavelengths. Across the broadband spectrum, our model maintains consistently low prediction error, reflecting its robustness in handling both trained and untrained wavelengths. In contrast, NeurOLight exhibits considerable performance fluctuations, with error spikes emerging in the middle of the untrained wavelength regions.

Figure~\ref{fig3}c, d and e, f present the results for the spectrum splitter (Fig.~\ref{fig2}d) and straight waveguide simulations (Fig.~\ref{fig2}g), respectively. Similar to the metalens case, NeurOLight accurately predicts electric fields at trained wavelengths but exhibits noticeable error peaks at untrained wavelengths. In contrast, our model exhibits a stable error profile across the entire wavelength spectrum, with minimal variation even at wavelengths unseen during training. Figure~\ref{fig3}d and f illustrate the results of the design region. Notably, our model consistently outperforms the compared baseline, demonstrating robustness against the continuously varying wavelengths that typically degrade the performance of other models.

Figure~\ref{fig4}a–c provide absolute error maps at 410nm, the wavelength where models exhibit the highest prediction errors. The baseline models, including NeurOLight, clearly show significant prediction errors. In contrast, our proposed model maintains uniformly low errors across the entire domain in all simulation scenarios. This result underscores our model’s robust capability to accurately capture both complex light-matter interaction in near- and far-field propagation characteristics at wavelengths unseen during training.

While the main text focuses on the comparison with NeurOLight, a complete comparison against all baseline models is included in Section 3 of the Supplementary Information to provide a broader context.

\subsection{Evaluating performance across varying dataset sizes}
\begin{figure}[htp]
\centering
\includegraphics[width=1\columnwidth]{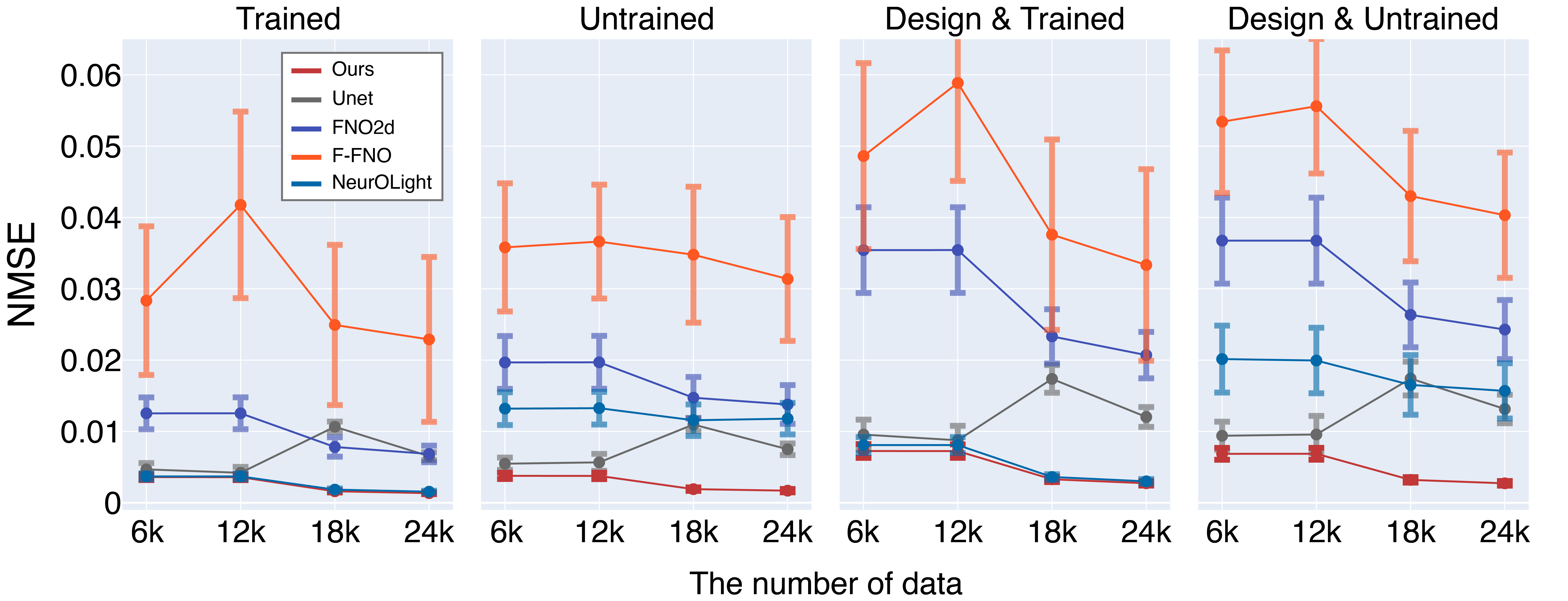}
\caption{Comparisons of field prediction accuracy across varying dataset sizes in the single layer metalens simulation. The ``Design'' label indicates evaluations in the design region, while results without this label correspond to the entire simulation domain. ``Trained'' and ``Untrained'' denote prediction performance for wavelengths observed during training and unseen (untrained) wavelengths, respectively. Additionally, the whiskers of the plots are standard deviations divided by 5.}
\label{fig5}
\end{figure}

To analyze the dependency between dataset size and model performance, we conduct additional experiments evaluating prediction accuracy across various training dataset sizes. As shown in Fig.~\ref{fig5}, our method consistently outperforms the baseline models for both trained and untrained wavelengths across all dataset sizes, including significantly reduced size (50\%). These results highlight the data efficiency of our approach. Such data-efficient performance has practical implications for applications where data collection is expensive or limited.

\subsection{Time and parameter efficiency}

\noindent
\setlength{\tabcolsep}{4pt} 
\renewcommand{\arraystretch}{1} 
\begin{table}[h]
\small
\centering
    \caption{Time efficiency of our method in comparison to a conventional numerical simulation (FDFD; Ceviche). The reported values represent the computational time per sample. The field prediction with our model was performed on a single RTX 6000 Ada GPU, while the numerical simulation was conducted on a 64-core AMD 5995WX CPU.\label{tab:timecomplexity}}
    \begin{tabular}{l|ccc}
        \toprule
        & Ours (batch size 1) & Ours (batch size 32) & Numerical Simulation\\
        \midrule
        Time (s) & 0.0172 & 0.0067 & 0.2786\\
        \bottomrule
    \end{tabular}
    \vspace{-0.3cm}
\end{table}

\begin{table}[ht]
\setlength{\tabcolsep}{24pt} 
\renewcommand{\arraystretch}{1} 
  \centering
  \caption{Comparison of the number of parameters across models.}
  \label{tab:param}
  \begin{tabular}{l l}
    \toprule
    Model & Parameters (M) \\
    \midrule
    U-Net             & 11.60 \\
    FNO               &  3.29 \\
    F-FNO             &  1.89 \\
    NeurOLight        &  1.65 \\
    \textbf{Ours}     &  \textbf{0.43} \\
    \bottomrule
  \end{tabular}
\end{table}

Our method also enables accelerated computation compared to a conventional numerical solver. To verify the time efficiency of our method, we conduct a computation time comparison between a conventional numerical simulation (FDFD; Ceviche~\cite{hughes2019forward}) and our model (Table~\ref{tab:timecomplexity}). Our model achieves a speedup of approximately 16× when using a batch size of 1 and up to 42× under the default batch size of 32, demonstrating an order-of-magnitude improvement in computation time efficiency relative to the numerical solver.

Beyond the time comparison, we also compare the parameter complexity of our model with baseline architectures. FNO~\cite{li2020fourier}, the foundation of our model, faces a significant challenge related to parameter complexity. For a 2D problem, each layer in FNO requires $C^2M_vM_h$ the number of parameters, where $M_v$ and $M_h$ denote the number of frequency modes in the vertical and horizontal directions, respectively, and $C$ represents the number of channels. This leads to substantial parameter numbers and can result in large networks.

In contrast, our surrogate solver uses only 0.43 million parameters, representing a 86.9\% reduction compared to FNO2d~\cite{li2020fourier}'s 3.29 million parameters and a 96.33\% reduction relative to UNet~\cite{ho2020denoising, gupta2022towards}’s 11.60 million parameters. This significant reduction arises from our Fourier Group Convolutional Shuffling (Section 2.1 in the Supplementary Information), which splits each Fourier layer into separate horizontal and vertical transforms and applies group-wise weight sharing with channel shuffling. This compresses the spectral weight tensor from $C^2M_vM_h$ to $C^2(M_v+M_h)/(4G)$. Despite these smaller parameter numbers, our model achieves state-of-the-art accuracy in both the design region and the propagation region.

\section{Discussion}\label{sec12}

In this work, we verify that the light-matter interaction in electromagnetic systems exhibits highly nonlinear and complex variations across continuously changing wavelengths. Consequently, predicting fields at unobserved intermediate wavelengths, which lie between the discrete wavelengths seen during training, remains a significant challenge.

In most cases, the condition parameters of physical phenomena modeled by PDEs are inherently continuous. Accordingly, a surrogate solver that serves as a reliable approximation should capture how physical responses vary across the continuous parameter space. However, it is critical that the learned model reflects not just the training cases but also the entire condition space including unseen cases during training, since training typically cover a finite set of discrete condition values. Therefore, we contend that a physically reliable surrogate solver should be able to generalize to unseen conditions lying between the training samples.
To achieve this, it is essential to ensure robust predictions not only for the parameter values observed during training but also for the parameter values unobserved during training.
A surrogate solver capable of accurately predicting across this continuous parameter space may provide a trustworthy alternative to traditional numerical solvers.
Motivated by the above claims, we determined to focus on exploring the generalization capability regarding limitedly observed condition values during training.

Our main findings are as follows: (i) Our method significantly improves predictive accuracy, especially at untrained wavelengths. This improvement addresses the inherent weaknesses of previous surrogate solvers in handling complex electromagnetic phenomena. (ii) our experiment on dataset size (Fig.~\ref{fig5}) indicates that our model maintains strong interpolation and field prediction accuracy even when the training dataset is reduced by up to 50\% (6k). This suggests that the proposed method holds promise for data-scarce scenarios and enables rapid prototyping and accelerated training processes. (iii) Our model also achieves time and parameter efficiency, providing over an order of magnitude speedup compared to conventional FDFD simulation~\cite{hughes2019forward} and requiring a significantly smaller number of parameters than FNO. Inference time per sample is reduced by approximately 16 times for single-batch evaluation and up to 42 times when leveraging inference with a batch size of 32 (Table~\ref{tab:timecomplexity}). Moreover, our model's parameter number shows a 86.9\% reduction relative to FNO2d~\cite{gu2022neurolight}'s 3.29 million parameters.

A core contribution is the introduction of Spectral Consistency and Wave-Informed element-wise Multiplicative Encoding (WIME). These methods explicitly concentrate on critical wavelength-dependent spatial frequency characteristics. By embedding physically meaningful wave priors and employing multiplicative integration, WIME robustly captures the intricate relationship between wavelength and spatial frequency distributions.

While a number of alternative conditioning strategies have been proposed in previous works, WIME differs fundamentally in both objective and design.
\begin{itemize}
    \item Refs.~\cite{mao2024towards,gupta2022towards} modulate the weights of Fourier layers using vector-encoded PDE parameters. Although such conditioning schemes are compatible with FNOs, they lack the ability to effectively capture complex electromagnetic phenomena across a wide range of physical parameters, a limitation we analyze in detail in Section 4.3 of the Supplementary Information.
    \item Channel-attention-based parameter embedding (CAPE)~\cite{takamoto2023learning} employs channel attention mechanisms based on vector-encoded PDE parameters. While this method has shown effectiveness in time-dependent PDE scenarios, it does not explicitly address how to spatially integrate conditional information.
    \item Finally, NeurOLight~\cite{gu2022neurolight} simply concatenates condition information along the input’s spatial dimensions.
\end{itemize}

On the other hand, our work focuses on effectively reflecting condition information in the spatial dimension and satisfying the introduced principle (spectral consistency), providing a new conditioning method with a different objective and significantly better generalization performance. This makes WIME not just a variant of existing schemes, but a fundamentally distinct and purpose-built conditioning method.

Despite these advances, several limitations and avenues for future work remain. First, in the high-\(Q\) regime or in designs with extreme permittivity contrast, the assumption of spectral consistency can break down, leading to modest degradation in prediction accuracy. Second, while our experiments focused on two-dimensional geometries under normal incidence, extending the approach to full three-dimensional devices, oblique incidence angles, and polarization diversity will be crucial for real-world applications in metasurface~\cite{yu2014flat, li20203d, lee2018metasurface} and photonics design~\cite{yang2024titanium, jang2025inverse, roques2022toward, fan20243d}.

\section{Methods}\label{sec11}

\textbf{Problem setting}
Let $\Omega \subset \mathbb{R}^d$,  $\mathcal{A} = \mathcal{A}(\Omega; \varepsilon^{d_a})$ where \(\varepsilon=\{\epsilon_{\text{air}}, \epsilon_{\text{material}}\}\), and $\mathcal{U}=\mathcal{U}(\Omega; \mathbb{C}^{d_u})$ be a bounded open set of the underlying domain, infinite-dimensional spaces of the relative permittivity and electric field of the simulation. $\epsilon_{\text{air}}$ and $\epsilon_{\text{material}}$ represent the relative permittivity values of air and material. Additionally, we assume that $\mathcal{W} \subset \mathbb{R}_{\geq 0}$ is the broadband range, such as the visible light spectrum, and $\tilde{\mathcal{W}} \subset \mathcal{W}$ is a discrete collection of wavelengths evenly distributed at a specific interval from $\mathcal{W}$. Our model $\mathcal G_\theta$ learns an ideal electromagnetic simulator $\mathcal G^{\dagger}:\mathcal{A} \rightarrow \mathcal{U}$ for continuous wavelengths $\mathcal{W}$ by mapping between infinite-dimensional function spaces using a finite set of Maxwell PDE input-output pairs $\{(w_j, a_j), u_j\}^N_{j=1}$, where $w \in \tilde{\mathcal{W}}$.

\textbf{Datasets}
We aim to predict the electric field for all wavelengths from discrete data in the 400-700nm (visible light) range. To achieve this, we generate field data for discrete wavelengths as our dataset. 
For each data, structures are randomly placed in the design region, and the field data resulting from FDFD simulations, along with the permittivity($\varepsilon$) of the materials, wavelength($\lambda$), and structural information, are used as pairs for training. The training set consists of 12,000 data sampled randomly at 20nm intervals within the 400-700nm wavelength range.
As mentioned above, our goal is to achieve seamless field prediction across the broadband wavelength. Therefore, to use unseen data for testing, we use 6,020 data each for the test and validation sets, composed of 20 data for each wavelength, sampled at 1nm intervals between 400 and 700nm. The dataset generation is conducted using a 64-core AMD 5995WX CPU. 

\textbf{Training Objective and Evaluation Metric} 
Fields typically exhibit different statistics despite the fixed source power. We employ the normalized mean squared error (N-MSE) objective, $L(\mathcal G_\theta(a), \mathcal G^\dagger(a)) = (||\mathcal G_\theta(a) - \mathcal G^\dagger(a)||^2_2 / ||\mathcal G^\dagger(a)||^2_2)$, to distribute the optimization effort evenly across several field data.

\textbf{Training Details}
Our model training runs for 200 epochs with 32 batch size. The best model is selected based on the lowest loss in predicting previously unseen wavelengths during the training phase. We employ the GELU (Gaussian Error Linear Unit)~\cite{hendrycks2016gaussian} activation function and the AdamW~\cite{loshchilov2017decoupled} optimizer with the following parameters: a learning rate of 0.002, beta values of (0.9, 0.999), epsilon set to $10^{-8}$, and a weight decay factor of 0.0001. 
To dynamically adjust the learning rate, we utilize a cosine annealing learning rate scheduler~\cite{loshchilov2016sgdr} with a minimum learning rate of 0.00001. Mode is set to (50, 60). We leverage PyTorch~\cite{paszke2019pytorch}, and the training process is conducted using a single RTX 6000 Ada.

\section{Data availability}
All data supporting the findings of this study are available within the article and its Supplementary Information. Source data are provided with this paper. And the photonic structure datasets are available at \url{https://drive.google.com/drive/folders/1YRuzbOaIv3q6AjBgvagha2H2cizaP1Rh?usp=sharing}.

\section{Code availability}
The code used for the experiments presented in this paper is available at \url{https://github.com/yhy258/WINO_interp}.

\bibliography{main-bib}


\begin{thebibliography}{71}
\ifx \bisbn   \undefined \def \bisbn  #1{ISBN #1}\fi
\ifx \binits  \undefined \def \binits#1{#1}\fi
\ifx \bauthor  \undefined \def \bauthor#1{#1}\fi
\ifx \batitle  \undefined \def \batitle#1{#1}\fi
\ifx \bjtitle  \undefined \def \bjtitle#1{#1}\fi
\ifx \bvolume  \undefined \def \bvolume#1{\textbf{#1}}\fi
\ifx \byear  \undefined \def \byear#1{#1}\fi
\ifx \bissue  \undefined \def \bissue#1{#1}\fi
\ifx \bfpage  \undefined \def \bfpage#1{#1}\fi
\ifx \blpage  \undefined \def \blpage #1{#1}\fi
\ifx \burl  \undefined \def \burl#1{\textsf{#1}}\fi
\ifx \doiurl  \undefined \def \doiurl#1{\url{https://doi.org/#1}}\fi
\ifx \betal  \undefined \def \betal{\textit{et al.}}\fi
\ifx \binstitute  \undefined \def \binstitute#1{#1}\fi
\ifx \binstitutionaled  \undefined \def \binstitutionaled#1{#1}\fi
\ifx \bctitle  \undefined \def \bctitle#1{#1}\fi
\ifx \beditor  \undefined \def \beditor#1{#1}\fi
\ifx \bpublisher  \undefined \def \bpublisher#1{#1}\fi
\ifx \bbtitle  \undefined \def \bbtitle#1{#1}\fi
\ifx \bedition  \undefined \def \bedition#1{#1}\fi
\ifx \bseriesno  \undefined \def \bseriesno#1{#1}\fi
\ifx \blocation  \undefined \def \blocation#1{#1}\fi
\ifx \bsertitle  \undefined \def \bsertitle#1{#1}\fi
\ifx \bsnm \undefined \def \bsnm#1{#1}\fi
\ifx \bsuffix \undefined \def \bsuffix#1{#1}\fi
\ifx \bparticle \undefined \def \bparticle#1{#1}\fi
\ifx \barticle \undefined \def \barticle#1{#1}\fi
\bibcommenthead
\ifx \bconfdate \undefined \def \bconfdate #1{#1}\fi
\ifx \botherref \undefined \def \botherref #1{#1}\fi
\ifx \url \undefined \def \url#1{\textsf{#1}}\fi
\ifx \bchapter \undefined \def \bchapter#1{#1}\fi
\ifx \bbook \undefined \def \bbook#1{#1}\fi
\ifx \bcomment \undefined \def \bcomment#1{#1}\fi
\ifx \oauthor \undefined \def \oauthor#1{#1}\fi
\ifx \citeauthoryear \undefined \def \citeauthoryear#1{#1}\fi
\ifx \endbibitem  \undefined \def \endbibitem {}\fi
\ifx \bconflocation  \undefined \def \bconflocation#1{#1}\fi
\ifx \arxivurl  \undefined \def \arxivurl#1{\textsf{#1}}\fi
\csname PreBibitemsHook\endcsname

\bibitem[\protect\citeauthoryear{Wen et~al.}{2019}]{wen2019progressive}
\begin{botherref}
\oauthor{\bsnm{Wen}, \binits{F.}},
\oauthor{\bsnm{Jiang}, \binits{J.}},
\oauthor{\bsnm{Fan}, \binits{J.A.}}:
Progressive-growing of generative adversarial networks for metasurface optimization.
arXiv preprint arXiv:1911.13029
(2019)
\end{botherref}
\endbibitem

\bibitem[\protect\citeauthoryear{Chen et~al.}{2022}]{chen2022high}
\begin{barticle}
\bauthor{\bsnm{Chen}, \binits{M.}},
\bauthor{\bsnm{Lupoiu}, \binits{R.}},
\bauthor{\bsnm{Mao}, \binits{C.}},
\bauthor{\bsnm{Huang}, \binits{D.-H.}},
\bauthor{\bsnm{Jiang}, \binits{J.}},
\bauthor{\bsnm{Lalanne}, \binits{P.}},
\bauthor{\bsnm{Fan}, \binits{J.A.}}:
\batitle{High speed simulation and freeform optimization of nanophotonic devices with physics-augmented deep learning}.
\bjtitle{ACS Photonics}
\bvolume{9}(\bissue{9}),
\bfpage{3110}--\blpage{3123}
(\byear{2022})
\end{barticle}
\endbibitem

\bibitem[\protect\citeauthoryear{Seo et~al.}{2025}]{seo2025physics}
\begin{botherref}
\oauthor{\bsnm{Seo}, \binits{D.}},
\oauthor{\bsnm{Um}, \binits{S.}},
\oauthor{\bsnm{Lee}, \binits{S.}},
\oauthor{\bsnm{Ye}, \binits{J.C.}},
\oauthor{\bsnm{Chung}, \binits{H.}}:
Physics-guided and fabrication-aware inverse design of photonic devices using diffusion models.
arXiv preprint arXiv:2504.17077
(2025)
\end{botherref}
\endbibitem

\bibitem[\protect\citeauthoryear{Seo et~al.}{2024}]{seo2024high}
\begin{bchapter}
\bauthor{\bsnm{Seo}, \binits{J.}},
\bauthor{\bsnm{Kang}, \binits{C.}},
\bauthor{\bsnm{Seo}, \binits{D.}},
\bauthor{\bsnm{Chung}, \binits{H.}}:
\bctitle{High-speed multiwavelength adjoint optimization with surrogate solver}.
In: \bbtitle{2024 Conference on Lasers and Electro-Optics Pacific Rim (CLEO-PR)},
pp. \bfpage{1}--\blpage{2}
(\byear{2024}).
\bcomment{IEEE}
\end{bchapter}
\endbibitem

\bibitem[\protect\citeauthoryear{Kang et~al.}{2025}]{kang2025adjoint}
\begin{botherref}
\oauthor{\bsnm{Kang}, \binits{C.}},
\oauthor{\bsnm{Seo}, \binits{J.}},
\oauthor{\bsnm{Jang}, \binits{I.}},
\oauthor{\bsnm{Chung}, \binits{H.}}:
Adjoint method-based fourier neural operator surrogate solver for wavefront shaping in tunable metasurfaces.
iScience
\textbf{28}(1)
(2025)
\end{botherref}
\endbibitem

\bibitem[\protect\citeauthoryear{Seo et~al.}{2023}]{seo2023deep}
\begin{botherref}
\oauthor{\bsnm{Seo}, \binits{J.}},
\oauthor{\bsnm{Jo}, \binits{J.}},
\oauthor{\bsnm{Kim}, \binits{J.}},
\oauthor{\bsnm{Kang}, \binits{J.}},
\oauthor{\bsnm{Kang}, \binits{C.}},
\oauthor{\bsnm{Moon}, \binits{S.}},
\oauthor{\bsnm{Lee}, \binits{E.}},
\oauthor{\bsnm{Hong}, \binits{J.}},
\oauthor{\bsnm{Rho}, \binits{J.}},
\oauthor{\bsnm{Chung}, \binits{H.}}:
Deep-learning-driven end-to-end metalens imaging.
arXiv preprint arXiv:2312.02669
(2023)
\end{botherref}
\endbibitem

\bibitem[\protect\citeauthoryear{Tseng et~al.}{2021}]{tseng2021neural}
\begin{barticle}
\bauthor{\bsnm{Tseng}, \binits{E.}},
\bauthor{\bsnm{Colburn}, \binits{S.}},
\bauthor{\bsnm{Whitehead}, \binits{J.}},
\bauthor{\bsnm{Huang}, \binits{L.}},
\bauthor{\bsnm{Baek}, \binits{S.-H.}},
\bauthor{\bsnm{Majumdar}, \binits{A.}},
\bauthor{\bsnm{Heide}, \binits{F.}}:
\batitle{Neural nano-optics for high-quality thin lens imaging}.
\bjtitle{Nature communications}
\bvolume{12}(\bissue{1}),
\bfpage{6493}
(\byear{2021})
\end{barticle}
\endbibitem

\bibitem[\protect\citeauthoryear{Deb et~al.}{2022}]{deb2022fouriernets}
\begin{barticle}
\bauthor{\bsnm{Deb}, \binits{D.}},
\bauthor{\bsnm{Jiao}, \binits{Z.}},
\bauthor{\bsnm{Sims}, \binits{R.}},
\bauthor{\bsnm{Chen}, \binits{A.}},
\bauthor{\bsnm{Broxton}, \binits{M.}},
\bauthor{\bsnm{Ahrens}, \binits{M.B.}},
\bauthor{\bsnm{Podgorski}, \binits{K.}},
\bauthor{\bsnm{Turaga}, \binits{S.C.}}:
\batitle{Fouriernets enable the design of highly non-local optical encoders for computational imaging}.
\bjtitle{Advances in Neural Information Processing Systems}
\bvolume{35},
\bfpage{25224}--\blpage{25236}
(\byear{2022})
\end{barticle}
\endbibitem

\bibitem[\protect\citeauthoryear{Tan et~al.}{2023}]{tan2023photonic}
\begin{barticle}
\bauthor{\bsnm{Tan}, \binits{M.}},
\bauthor{\bsnm{Xu}, \binits{X.}},
\bauthor{\bsnm{Boes}, \binits{A.}},
\bauthor{\bsnm{Corcoran}, \binits{B.}},
\bauthor{\bsnm{Nguyen}, \binits{T.G.}},
\bauthor{\bsnm{Chu}, \binits{S.T.}},
\bauthor{\bsnm{Little}, \binits{B.E.}},
\bauthor{\bsnm{Morandotti}, \binits{R.}},
\bauthor{\bsnm{Wu}, \binits{J.}},
\bauthor{\bsnm{Mitchell}, \binits{A.}}, \betal:
\batitle{Photonic signal processor based on a kerr microcomb for real-time video image processing}.
\bjtitle{Communications Engineering}
\bvolume{2}(\bissue{1}),
\bfpage{94}
(\byear{2023})
\end{barticle}
\endbibitem

\bibitem[\protect\citeauthoryear{Fan et~al.}{2020}]{fan2020advancing}
\begin{barticle}
\bauthor{\bsnm{Fan}, \binits{Q.}},
\bauthor{\bsnm{Zhou}, \binits{G.}},
\bauthor{\bsnm{Gui}, \binits{T.}},
\bauthor{\bsnm{Lu}, \binits{C.}},
\bauthor{\bsnm{Lau}, \binits{A.P.T.}}:
\batitle{Advancing theoretical understanding and practical performance of signal processing for nonlinear optical communications through machine learning}.
\bjtitle{Nature Communications}
\bvolume{11}(\bissue{1}),
\bfpage{3694}
(\byear{2020})
\end{barticle}
\endbibitem

\bibitem[\protect\citeauthoryear{Huang et~al.}{2021}]{huang2021all}
\begin{barticle}
\bauthor{\bsnm{Huang}, \binits{Z.}},
\bauthor{\bsnm{Wang}, \binits{P.}},
\bauthor{\bsnm{Liu}, \binits{J.}},
\bauthor{\bsnm{Xiong}, \binits{W.}},
\bauthor{\bsnm{He}, \binits{Y.}},
\bauthor{\bsnm{Xiao}, \binits{J.}},
\bauthor{\bsnm{Ye}, \binits{H.}},
\bauthor{\bsnm{Li}, \binits{Y.}},
\bauthor{\bsnm{Chen}, \binits{S.}},
\bauthor{\bsnm{Fan}, \binits{D.}}:
\batitle{All-optical signal processing of vortex beams with diffractive deep neural networks}.
\bjtitle{Physical Review Applied}
\bvolume{15}(\bissue{1}),
\bfpage{014037}
(\byear{2021})
\end{barticle}
\endbibitem

\bibitem[\protect\citeauthoryear{Lin et~al.}{2018}]{lin2018all}
\begin{barticle}
\bauthor{\bsnm{Lin}, \binits{X.}},
\bauthor{\bsnm{Rivenson}, \binits{Y.}},
\bauthor{\bsnm{Yardimci}, \binits{N.T.}},
\bauthor{\bsnm{Veli}, \binits{M.}},
\bauthor{\bsnm{Luo}, \binits{Y.}},
\bauthor{\bsnm{Jarrahi}, \binits{M.}},
\bauthor{\bsnm{Ozcan}, \binits{A.}}:
\batitle{All-optical machine learning using diffractive deep neural networks}.
\bjtitle{Science}
\bvolume{361}(\bissue{6406}),
\bfpage{1004}--\blpage{1008}
(\byear{2018})
\end{barticle}
\endbibitem

\bibitem[\protect\citeauthoryear{Xu et~al.}{2021}]{xu202111}
\begin{barticle}
\bauthor{\bsnm{Xu}, \binits{X.}},
\bauthor{\bsnm{Tan}, \binits{M.}},
\bauthor{\bsnm{Corcoran}, \binits{B.}},
\bauthor{\bsnm{Wu}, \binits{J.}},
\bauthor{\bsnm{Boes}, \binits{A.}},
\bauthor{\bsnm{Nguyen}, \binits{T.G.}},
\bauthor{\bsnm{Chu}, \binits{S.T.}},
\bauthor{\bsnm{Little}, \binits{B.E.}},
\bauthor{\bsnm{Hicks}, \binits{D.G.}},
\bauthor{\bsnm{Morandotti}, \binits{R.}}, \betal:
\batitle{11 tops photonic convolutional accelerator for optical neural networks}.
\bjtitle{Nature}
\bvolume{589}(\bissue{7840}),
\bfpage{44}--\blpage{51}
(\byear{2021})
\end{barticle}
\endbibitem

\bibitem[\protect\citeauthoryear{Fu et~al.}{2024}]{fu2024optical}
\begin{barticle}
\bauthor{\bsnm{Fu}, \binits{T.}},
\bauthor{\bsnm{Zhang}, \binits{J.}},
\bauthor{\bsnm{Sun}, \binits{R.}},
\bauthor{\bsnm{Huang}, \binits{Y.}},
\bauthor{\bsnm{Xu}, \binits{W.}},
\bauthor{\bsnm{Yang}, \binits{S.}},
\bauthor{\bsnm{Zhu}, \binits{Z.}},
\bauthor{\bsnm{Chen}, \binits{H.}}:
\batitle{Optical neural networks: progress and challenges}.
\bjtitle{Light: Science \& Applications}
\bvolume{13}(\bissue{1}),
\bfpage{263}
(\byear{2024})
\end{barticle}
\endbibitem

\bibitem[\protect\citeauthoryear{d’Arco et~al.}{2022}]{d2022physics}
\begin{barticle}
\bauthor{\bsnm{d’Arco}, \binits{A.}},
\bauthor{\bsnm{Xia}, \binits{F.}},
\bauthor{\bsnm{Boniface}, \binits{A.}},
\bauthor{\bsnm{Dong}, \binits{J.}},
\bauthor{\bsnm{Gigan}, \binits{S.}}:
\batitle{Physics-based neural network for non-invasive control of coherent light in scattering media}.
\bjtitle{Optics express}
\bvolume{30}(\bissue{17}),
\bfpage{30845}--\blpage{30856}
(\byear{2022})
\end{barticle}
\endbibitem

\bibitem[\protect\citeauthoryear{Wright et~al.}{2022}]{wright2022deep}
\begin{barticle}
\bauthor{\bsnm{Wright}, \binits{L.G.}},
\bauthor{\bsnm{Onodera}, \binits{T.}},
\bauthor{\bsnm{Stein}, \binits{M.M.}},
\bauthor{\bsnm{Wang}, \binits{T.}},
\bauthor{\bsnm{Schachter}, \binits{D.T.}},
\bauthor{\bsnm{Hu}, \binits{Z.}},
\bauthor{\bsnm{McMahon}, \binits{P.L.}}:
\batitle{Deep physical neural networks trained with backpropagation}.
\bjtitle{Nature}
\bvolume{601}(\bissue{7894}),
\bfpage{549}--\blpage{555}
(\byear{2022})
\end{barticle}
\endbibitem

\bibitem[\protect\citeauthoryear{Momeni et~al.}{2023}]{momeni2023backpropagation}
\begin{barticle}
\bauthor{\bsnm{Momeni}, \binits{A.}},
\bauthor{\bsnm{Rahmani}, \binits{B.}},
\bauthor{\bsnm{Mall{\'e}jac}, \binits{M.}},
\bauthor{\bsnm{Del~Hougne}, \binits{P.}},
\bauthor{\bsnm{Fleury}, \binits{R.}}:
\batitle{Backpropagation-free training of deep physical neural networks}.
\bjtitle{Science}
\bvolume{382}(\bissue{6676}),
\bfpage{1297}--\blpage{1303}
(\byear{2023})
\end{barticle}
\endbibitem

\bibitem[\protect\citeauthoryear{Jiang et~al.}{2021}]{jiang2021deep}
\begin{barticle}
\bauthor{\bsnm{Jiang}, \binits{J.}},
\bauthor{\bsnm{Chen}, \binits{M.}},
\bauthor{\bsnm{Fan}, \binits{J.A.}}:
\batitle{Deep neural networks for the evaluation and design of photonic devices}.
\bjtitle{Nature Reviews Materials}
\bvolume{6}(\bissue{8}),
\bfpage{679}--\blpage{700}
(\byear{2021})
\end{barticle}
\endbibitem

\bibitem[\protect\citeauthoryear{So et~al.}{2020}]{so2020deep}
\begin{barticle}
\bauthor{\bsnm{So}, \binits{S.}},
\bauthor{\bsnm{Badloe}, \binits{T.}},
\bauthor{\bsnm{Noh}, \binits{J.}},
\bauthor{\bsnm{Bravo-Abad}, \binits{J.}},
\bauthor{\bsnm{Rho}, \binits{J.}}:
\batitle{Deep learning enabled inverse design in nanophotonics}.
\bjtitle{Nanophotonics}
\bvolume{9}(\bissue{5}),
\bfpage{1041}--\blpage{1057}
(\byear{2020})
\end{barticle}
\endbibitem

\bibitem[\protect\citeauthoryear{Gu et~al.}{2022}]{gu2022neurolight}
\begin{barticle}
\bauthor{\bsnm{Gu}, \binits{J.}},
\bauthor{\bsnm{Gao}, \binits{Z.}},
\bauthor{\bsnm{Feng}, \binits{C.}},
\bauthor{\bsnm{Zhu}, \binits{H.}},
\bauthor{\bsnm{Chen}, \binits{R.}},
\bauthor{\bsnm{Boning}, \binits{D.}},
\bauthor{\bsnm{Pan}, \binits{D.}}:
\batitle{Neurolight: A physics-agnostic neural operator enabling parametric photonic device simulation}.
\bjtitle{Advances in Neural Information Processing Systems}
\bvolume{35},
\bfpage{14623}--\blpage{14636}
(\byear{2022})
\end{barticle}
\endbibitem

\bibitem[\protect\citeauthoryear{Malkiel et~al.}{2018}]{malkiel2018plasmonic}
\begin{barticle}
\bauthor{\bsnm{Malkiel}, \binits{I.}},
\bauthor{\bsnm{Mrejen}, \binits{M.}},
\bauthor{\bsnm{Nagler}, \binits{A.}},
\bauthor{\bsnm{Arieli}, \binits{U.}},
\bauthor{\bsnm{Wolf}, \binits{L.}},
\bauthor{\bsnm{Suchowski}, \binits{H.}}:
\batitle{Plasmonic nanostructure design and characterization via deep learning}.
\bjtitle{Light: Science \& Applications}
\bvolume{7}(\bissue{1}),
\bfpage{60}
(\byear{2018})
\end{barticle}
\endbibitem

\bibitem[\protect\citeauthoryear{Peurifoy et~al.}{2018}]{peurifoy2018nanophotonic}
\begin{barticle}
\bauthor{\bsnm{Peurifoy}, \binits{J.}},
\bauthor{\bsnm{Shen}, \binits{Y.}},
\bauthor{\bsnm{Jing}, \binits{L.}},
\bauthor{\bsnm{Yang}, \binits{Y.}},
\bauthor{\bsnm{Cano-Renteria}, \binits{F.}},
\bauthor{\bsnm{DeLacy}, \binits{B.G.}},
\bauthor{\bsnm{Joannopoulos}, \binits{J.D.}},
\bauthor{\bsnm{Tegmark}, \binits{M.}},
\bauthor{\bsnm{Solja{\v{c}}i{\'c}}, \binits{M.}}:
\batitle{Nanophotonic particle simulation and inverse design using artificial neural networks}.
\bjtitle{Science advances}
\bvolume{4}(\bissue{6}),
\bfpage{4206}
(\byear{2018})
\end{barticle}
\endbibitem

\bibitem[\protect\citeauthoryear{Sajedian et~al.}{2019}]{sajedian2019finding}
\begin{barticle}
\bauthor{\bsnm{Sajedian}, \binits{I.}},
\bauthor{\bsnm{Kim}, \binits{J.}},
\bauthor{\bsnm{Rho}, \binits{J.}}:
\batitle{Finding the optical properties of plasmonic structures by image processing using a combination of convolutional neural networks and recurrent neural networks}.
\bjtitle{Microsystems \& nanoengineering}
\bvolume{5}(\bissue{1}),
\bfpage{27}
(\byear{2019})
\end{barticle}
\endbibitem

\bibitem[\protect\citeauthoryear{Lu et~al.}{2021}]{lu2021physics}
\begin{barticle}
\bauthor{\bsnm{Lu}, \binits{L.}},
\bauthor{\bsnm{Pestourie}, \binits{R.}},
\bauthor{\bsnm{Yao}, \binits{W.}},
\bauthor{\bsnm{Wang}, \binits{Z.}},
\bauthor{\bsnm{Verdugo}, \binits{F.}},
\bauthor{\bsnm{Johnson}, \binits{S.G.}}:
\batitle{Physics-informed neural networks with hard constraints for inverse design}.
\bjtitle{SIAM Journal on Scientific Computing}
\bvolume{43}(\bissue{6}),
\bfpage{1105}--\blpage{1132}
(\byear{2021})
\end{barticle}
\endbibitem

\bibitem[\protect\citeauthoryear{Zhelyeznyakov et~al.}{2023}]{zhelyeznyakov2023large}
\begin{barticle}
\bauthor{\bsnm{Zhelyeznyakov}, \binits{M.}},
\bauthor{\bsnm{Fr{\"o}ch}, \binits{J.}},
\bauthor{\bsnm{Wirth-Singh}, \binits{A.}},
\bauthor{\bsnm{Noh}, \binits{J.}},
\bauthor{\bsnm{Rho}, \binits{J.}},
\bauthor{\bsnm{Brunton}, \binits{S.}},
\bauthor{\bsnm{Majumdar}, \binits{A.}}:
\batitle{Large area optimization of meta-lens via data-free machine learning}.
\bjtitle{Communications Engineering}
\bvolume{2}(\bissue{1}),
\bfpage{60}
(\byear{2023})
\end{barticle}
\endbibitem

\bibitem[\protect\citeauthoryear{Piao et~al.}{2024}]{piao2024domain}
\begin{botherref}
\oauthor{\bsnm{Piao}, \binits{S.}},
\oauthor{\bsnm{Gu}, \binits{H.}},
\oauthor{\bsnm{Wang}, \binits{A.}},
\oauthor{\bsnm{Qin}, \binits{P.}}:
A domain-adaptive physics-informed neural network for inverse problems of maxwell's equations in heterogeneous media.
IEEE Antennas and Wireless Propagation Letters
(2024)
\end{botherref}
\endbibitem

\bibitem[\protect\citeauthoryear{Lim and Psaltis}{2022}]{lim2022maxwellnet}
\begin{botherref}
\oauthor{\bsnm{Lim}, \binits{J.}},
\oauthor{\bsnm{Psaltis}, \binits{D.}}:
Maxwellnet: Physics-driven deep neural network training based on maxwell’s equations.
Apl Photonics
\textbf{7}(1)
(2022)
\end{botherref}
\endbibitem

\bibitem[\protect\citeauthoryear{Kovachki et~al.}{2023}]{kovachki2023neural}
\begin{barticle}
\bauthor{\bsnm{Kovachki}, \binits{N.}},
\bauthor{\bsnm{Li}, \binits{Z.}},
\bauthor{\bsnm{Liu}, \binits{B.}},
\bauthor{\bsnm{Azizzadenesheli}, \binits{K.}},
\bauthor{\bsnm{Bhattacharya}, \binits{K.}},
\bauthor{\bsnm{Stuart}, \binits{A.}},
\bauthor{\bsnm{Anandkumar}, \binits{A.}}:
\batitle{Neural operator: Learning maps between function spaces with applications to pdes}.
\bjtitle{Journal of Machine Learning Research}
\bvolume{24}(\bissue{89}),
\bfpage{1}--\blpage{97}
(\byear{2023})
\end{barticle}
\endbibitem

\bibitem[\protect\citeauthoryear{Azizzadenesheli et~al.}{2024}]{azizzadenesheli2024neural}
\begin{botherref}
\oauthor{\bsnm{Azizzadenesheli}, \binits{K.}},
\oauthor{\bsnm{Kovachki}, \binits{N.}},
\oauthor{\bsnm{Li}, \binits{Z.}},
\oauthor{\bsnm{Liu-Schiaffini}, \binits{M.}},
\oauthor{\bsnm{Kossaifi}, \binits{J.}},
\oauthor{\bsnm{Anandkumar}, \binits{A.}}:
Neural operators for accelerating scientific simulations and design.
Nature Reviews Physics,
1--9
(2024)
\end{botherref}
\endbibitem

\bibitem[\protect\citeauthoryear{Cao}{2021}]{cao2021choose}
\begin{barticle}
\bauthor{\bsnm{Cao}, \binits{S.}}:
\batitle{Choose a transformer: Fourier or galerkin}.
\bjtitle{Advances in neural information processing systems}
\bvolume{34},
\bfpage{24924}--\blpage{24940}
(\byear{2021})
\end{barticle}
\endbibitem

\bibitem[\protect\citeauthoryear{Hao et~al.}{2023}]{hao2023gnot}
\begin{bchapter}
\bauthor{\bsnm{Hao}, \binits{Z.}},
\bauthor{\bsnm{Wang}, \binits{Z.}},
\bauthor{\bsnm{Su}, \binits{H.}},
\bauthor{\bsnm{Ying}, \binits{C.}},
\bauthor{\bsnm{Dong}, \binits{Y.}},
\bauthor{\bsnm{Liu}, \binits{S.}},
\bauthor{\bsnm{Cheng}, \binits{Z.}},
\bauthor{\bsnm{Song}, \binits{J.}},
\bauthor{\bsnm{Zhu}, \binits{J.}}:
\bctitle{Gnot: A general neural operator transformer for operator learning}.
In: \bbtitle{International Conference on Machine Learning},
pp. \bfpage{12556}--\blpage{12569}
(\byear{2023}).
\bcomment{PMLR}
\end{bchapter}
\endbibitem

\bibitem[\protect\citeauthoryear{Li et~al.}{2022}]{li2022transformer}
\begin{botherref}
\oauthor{\bsnm{Li}, \binits{Z.}},
\oauthor{\bsnm{Meidani}, \binits{K.}},
\oauthor{\bsnm{Farimani}, \binits{A.B.}}:
Transformer for partial differential equations' operator learning.
arXiv preprint arXiv:2205.13671
(2022)
\end{botherref}
\endbibitem

\bibitem[\protect\citeauthoryear{Li et~al.}{2024}]{li2024scalable}
\begin{botherref}
\oauthor{\bsnm{Li}, \binits{Z.}},
\oauthor{\bsnm{Shu}, \binits{D.}},
\oauthor{\bsnm{Barati~Farimani}, \binits{A.}}:
Scalable transformer for pde surrogate modeling.
Advances in Neural Information Processing Systems
\textbf{36}
(2024)
\end{botherref}
\endbibitem

\bibitem[\protect\citeauthoryear{Raonic et~al.}{2023}]{raonic2023convolutional}
\begin{bchapter}
\bauthor{\bsnm{Raonic}, \binits{B.}},
\bauthor{\bsnm{Molinaro}, \binits{R.}},
\bauthor{\bsnm{Rohner}, \binits{T.}},
\bauthor{\bsnm{Mishra}, \binits{S.}},
\bauthor{\bsnm{Bezenac}, \binits{E.}}:
\bctitle{Convolutional neural operators}.
In: \bbtitle{ICLR 2023 Workshop on Physics for Machine Learning}
(\byear{2023})
\end{bchapter}
\endbibitem

\bibitem[\protect\citeauthoryear{Li et~al.}{2020}]{li2020fourier}
\begin{botherref}
\oauthor{\bsnm{Li}, \binits{Z.}},
\oauthor{\bsnm{Kovachki}, \binits{N.}},
\oauthor{\bsnm{Azizzadenesheli}, \binits{K.}},
\oauthor{\bsnm{Liu}, \binits{B.}},
\oauthor{\bsnm{Bhattacharya}, \binits{K.}},
\oauthor{\bsnm{Stuart}, \binits{A.}},
\oauthor{\bsnm{Anandkumar}, \binits{A.}}:
Fourier neural operator for parametric partial differential equations.
arXiv preprint arXiv:2010.08895
(2020)
\end{botherref}
\endbibitem

\bibitem[\protect\citeauthoryear{Tran et~al.}{2021}]{tran2021factorized}
\begin{botherref}
\oauthor{\bsnm{Tran}, \binits{A.}},
\oauthor{\bsnm{Mathews}, \binits{A.}},
\oauthor{\bsnm{Xie}, \binits{L.}},
\oauthor{\bsnm{Ong}, \binits{C.S.}}:
Factorized fourier neural operators.
arXiv preprint arXiv:2111.13802
(2021)
\end{botherref}
\endbibitem

\bibitem[\protect\citeauthoryear{Li et~al.}{2023}]{li2023fourier}
\begin{barticle}
\bauthor{\bsnm{Li}, \binits{Z.}},
\bauthor{\bsnm{Huang}, \binits{D.Z.}},
\bauthor{\bsnm{Liu}, \binits{B.}},
\bauthor{\bsnm{Anandkumar}, \binits{A.}}:
\batitle{Fourier neural operator with learned deformations for pdes on general geometries}.
\bjtitle{Journal of Machine Learning Research}
\bvolume{24}(\bissue{388}),
\bfpage{1}--\blpage{26}
(\byear{2023})
\end{barticle}
\endbibitem

\bibitem[\protect\citeauthoryear{Li et~al.}{2024}]{li2024geometry}
\begin{botherref}
\oauthor{\bsnm{Li}, \binits{Z.}},
\oauthor{\bsnm{Kovachki}, \binits{N.}},
\oauthor{\bsnm{Choy}, \binits{C.}},
\oauthor{\bsnm{Li}, \binits{B.}},
\oauthor{\bsnm{Kossaifi}, \binits{J.}},
\oauthor{\bsnm{Otta}, \binits{S.}},
\oauthor{\bsnm{Nabian}, \binits{M.A.}},
\oauthor{\bsnm{Stadler}, \binits{M.}},
\oauthor{\bsnm{Hundt}, \binits{C.}},
\oauthor{\bsnm{Azizzadenesheli}, \binits{K.}}, et al.:
Geometry-informed neural operator for large-scale 3d pdes.
Advances in Neural Information Processing Systems
\textbf{36}
(2024)
\end{botherref}
\endbibitem

\bibitem[\protect\citeauthoryear{Augenstein et~al.}{2023}]{augenstein2023neural}
\begin{barticle}
\bauthor{\bsnm{Augenstein}, \binits{Y.}},
\bauthor{\bsnm{Repan}, \binits{T.}},
\bauthor{\bsnm{Rockstuhl}, \binits{C.}}:
\batitle{Neural operator-based surrogate solver for free-form electromagnetic inverse design}.
\bjtitle{ACS Photonics}
\bvolume{10}(\bissue{5}),
\bfpage{1547}--\blpage{1557}
(\byear{2023})
\end{barticle}
\endbibitem

\bibitem[\protect\citeauthoryear{Mao et~al.}{2024}]{mao2024towards}
\begin{botherref}
\oauthor{\bsnm{Mao}, \binits{C.}},
\oauthor{\bsnm{Lupoiu}, \binits{R.}},
\oauthor{\bsnm{Dai}, \binits{T.}},
\oauthor{\bsnm{Chen}, \binits{M.}},
\oauthor{\bsnm{Fan}, \binits{J.A.}}:
Towards general neural surrogate solvers with specialized neural accelerators.
arXiv preprint arXiv:2405.02351
(2024)
\end{botherref}
\endbibitem

\bibitem[\protect\citeauthoryear{Zhang and Miller}{2020}]{zhang2020quasinormal}
\begin{botherref}
\oauthor{\bsnm{Zhang}, \binits{H.}},
\oauthor{\bsnm{Miller}, \binits{O.D.}}:
Quasinormal coupled mode theory.
arXiv preprint arXiv:2010.08650
(2020)
\end{botherref}
\endbibitem

\bibitem[\protect\citeauthoryear{Ruan and Fan}{2010}]{ruan2010superscattering}
\begin{barticle}
\bauthor{\bsnm{Ruan}, \binits{Z.}},
\bauthor{\bsnm{Fan}, \binits{S.}}:
\batitle{Superscattering of light from subwavelength nanostructures}.
\bjtitle{Physical review letters}
\bvolume{105}(\bissue{1}),
\bfpage{013901}
(\byear{2010})
\end{barticle}
\endbibitem

\bibitem[\protect\citeauthoryear{Goodman}{2005}]{goodman2005introduction}
\begin{bbook}
\bauthor{\bsnm{Goodman}, \binits{J.W.}}:
\bbtitle{Introduction to Fourier Optics}.
\bpublisher{Roberts and Company publishers}, \blocation{???}
(\byear{2005})
\end{bbook}
\endbibitem

\bibitem[\protect\citeauthoryear{Meade et~al.}{2008}]{meade2008photonic}
\begin{botherref}
\oauthor{\bsnm{Meade}, \binits{R.D.V.}},
\oauthor{\bsnm{Johnson}, \binits{S.G.}},
\oauthor{\bsnm{Winn}, \binits{J.N.}}:
Photonic crystals: Molding the flow of light.
Princeton University Press
(2008)
\end{botherref}
\endbibitem

\bibitem[\protect\citeauthoryear{Herzig~Sheinfux et~al.}{2016}]{herzig2016interplay}
\begin{barticle}
\bauthor{\bsnm{Herzig~Sheinfux}, \binits{H.}},
\bauthor{\bsnm{Kaminer}, \binits{I.}},
\bauthor{\bsnm{Genack}, \binits{A.Z.}},
\bauthor{\bsnm{Segev}, \binits{M.}}:
\batitle{Interplay between evanescence and disorder in deep subwavelength photonic structures}.
\bjtitle{Nature communications}
\bvolume{7}(\bissue{1}),
\bfpage{12927}
(\byear{2016})
\end{barticle}
\endbibitem

\bibitem[\protect\citeauthoryear{Lalanne and Lemercier-Lalanne}{1996}]{lalanne1996effective}
\begin{barticle}
\bauthor{\bsnm{Lalanne}, \binits{P.}},
\bauthor{\bsnm{Lemercier-Lalanne}, \binits{D.}}:
\batitle{On the effective medium theory of subwavelength periodic structures}.
\bjtitle{Journal of Modern Optics}
\bvolume{43}(\bissue{10}),
\bfpage{2063}--\blpage{2085}
(\byear{1996})
\end{barticle}
\endbibitem

\bibitem[\protect\citeauthoryear{Guenther}{2018}]{guenther2018modern}
\begin{bbook}
\bauthor{\bsnm{Guenther}, \binits{B.D.}}:
\bbtitle{Modern Optics}.
\bpublisher{Oxford University Press}, \blocation{???}
(\byear{2018})
\end{bbook}
\endbibitem

\bibitem[\protect\citeauthoryear{Park et~al.}{2019}]{park2019all}
\begin{barticle}
\bauthor{\bsnm{Park}, \binits{J.-S.}},
\bauthor{\bsnm{Zhang}, \binits{S.}},
\bauthor{\bsnm{She}, \binits{A.}},
\bauthor{\bsnm{Chen}, \binits{W.T.}},
\bauthor{\bsnm{Lin}, \binits{P.}},
\bauthor{\bsnm{Yousef}, \binits{K.M.}},
\bauthor{\bsnm{Cheng}, \binits{J.-X.}},
\bauthor{\bsnm{Capasso}, \binits{F.}}:
\batitle{All-glass, large metalens at visible wavelength using deep-ultraviolet projection lithography}.
\bjtitle{Nano letters}
\bvolume{19}(\bissue{12}),
\bfpage{8673}--\blpage{8682}
(\byear{2019})
\end{barticle}
\endbibitem

\bibitem[\protect\citeauthoryear{Chung and Miller}{2020}]{chung2020high}
\begin{barticle}
\bauthor{\bsnm{Chung}, \binits{H.}},
\bauthor{\bsnm{Miller}, \binits{O.D.}}:
\batitle{High-na achromatic metalenses by inverse design}.
\bjtitle{Optics Express}
\bvolume{28}(\bissue{5}),
\bfpage{6945}--\blpage{6965}
(\byear{2020})
\end{barticle}
\endbibitem

\bibitem[\protect\citeauthoryear{Bayati et~al.}{2020}]{bayati2020inverse}
\begin{barticle}
\bauthor{\bsnm{Bayati}, \binits{E.}},
\bauthor{\bsnm{Pestourie}, \binits{R.}},
\bauthor{\bsnm{Colburn}, \binits{S.}},
\bauthor{\bsnm{Lin}, \binits{Z.}},
\bauthor{\bsnm{Johnson}, \binits{S.G.}},
\bauthor{\bsnm{Majumdar}, \binits{A.}}:
\batitle{Inverse designed metalenses with extended depth of focus}.
\bjtitle{ACS photonics}
\bvolume{7}(\bissue{4}),
\bfpage{873}--\blpage{878}
(\byear{2020})
\end{barticle}
\endbibitem

\bibitem[\protect\citeauthoryear{Zou et~al.}{2022}]{zou2022pixel}
\begin{barticle}
\bauthor{\bsnm{Zou}, \binits{X.}},
\bauthor{\bsnm{Zhang}, \binits{Y.}},
\bauthor{\bsnm{Lin}, \binits{R.}},
\bauthor{\bsnm{Gong}, \binits{G.}},
\bauthor{\bsnm{Wang}, \binits{S.}},
\bauthor{\bsnm{Zhu}, \binits{S.}},
\bauthor{\bsnm{Wang}, \binits{Z.}}:
\batitle{Pixel-level bayer-type colour router based on metasurfaces}.
\bjtitle{Nature Communications}
\bvolume{13}(\bissue{1}),
\bfpage{3288}
(\byear{2022})
\end{barticle}
\endbibitem

\bibitem[\protect\citeauthoryear{Lee et~al.}{2024}]{lee2024inverse}
\begin{barticle}
\bauthor{\bsnm{Lee}, \binits{S.}},
\bauthor{\bsnm{Hong}, \binits{J.}},
\bauthor{\bsnm{Kang}, \binits{J.}},
\bauthor{\bsnm{Park}, \binits{J.}},
\bauthor{\bsnm{Lim}, \binits{J.}},
\bauthor{\bsnm{Lee}, \binits{T.}},
\bauthor{\bsnm{Jang}, \binits{M.S.}},
\bauthor{\bsnm{Chung}, \binits{H.}}:
\batitle{Inverse design of color routers in cmos image sensors: toward minimizing interpixel crosstalk}.
\bjtitle{Nanophotonics}
\bvolume{13}(\bissue{20}),
\bfpage{3895}--\blpage{3914}
(\byear{2024})
\end{barticle}
\endbibitem

\bibitem[\protect\citeauthoryear{Catrysse et~al.}{2022}]{catrysse2022subwavelength}
\begin{barticle}
\bauthor{\bsnm{Catrysse}, \binits{P.B.}},
\bauthor{\bsnm{Zhao}, \binits{N.}},
\bauthor{\bsnm{Jin}, \binits{W.}},
\bauthor{\bsnm{Fan}, \binits{S.}}:
\batitle{Subwavelength bayer rgb color routers with perfect optical efficiency}.
\bjtitle{Nanophotonics}
\bvolume{11}(\bissue{10}),
\bfpage{2381}--\blpage{2387}
(\byear{2022})
\end{barticle}
\endbibitem

\bibitem[\protect\citeauthoryear{Piggott et~al.}{2015}]{piggott2015inverse}
\begin{barticle}
\bauthor{\bsnm{Piggott}, \binits{A.Y.}},
\bauthor{\bsnm{Lu}, \binits{J.}},
\bauthor{\bsnm{Lagoudakis}, \binits{K.G.}},
\bauthor{\bsnm{Petykiewicz}, \binits{J.}},
\bauthor{\bsnm{Babinec}, \binits{T.M.}},
\bauthor{\bsnm{Vu{\v{c}}kovi{\'c}}, \binits{J.}}:
\batitle{Inverse design and demonstration of a compact and broadband on-chip wavelength demultiplexer}.
\bjtitle{Nature photonics}
\bvolume{9}(\bissue{6}),
\bfpage{374}--\blpage{377}
(\byear{2015})
\end{barticle}
\endbibitem

\bibitem[\protect\citeauthoryear{Bae et~al.}{2024}]{bae2024topology}
\begin{bchapter}
\bauthor{\bsnm{Bae}, \binits{M.}},
\bauthor{\bsnm{Kim}, \binits{C.}},
\bauthor{\bsnm{Lee}, \binits{S.}},
\bauthor{\bsnm{Choi}, \binits{M.}},
\bauthor{\bsnm{Lee}, \binits{M.}},
\bauthor{\bsnm{Jung}, \binits{H.}},
\bauthor{\bsnm{Kwon}, \binits{H.}},
\bauthor{\bsnm{Chung}, \binits{H.}}:
\bctitle{Topology optimization of lithium niobate mode converter}.
In: \bbtitle{Conference on Lasers and Electro-Optics/Pacific Rim},
pp. \bfpage{2}--\blpage{2}
(\byear{2024}).
\bcomment{Optica Publishing Group}
\end{bchapter}
\endbibitem

\bibitem[\protect\citeauthoryear{Chen et~al.}{2024}]{chen2024freeform}
\begin{barticle}
\bauthor{\bsnm{Chen}, \binits{T.}},
\bauthor{\bsnm{Xu}, \binits{M.}},
\bauthor{\bsnm{Pu}, \binits{M.}},
\bauthor{\bsnm{Zeng}, \binits{Q.}},
\bauthor{\bsnm{Zheng}, \binits{Y.}},
\bauthor{\bsnm{Xiao}, \binits{Y.}},
\bauthor{\bsnm{Ha}, \binits{Y.}},
\bauthor{\bsnm{Guo}, \binits{Y.}},
\bauthor{\bsnm{Zhang}, \binits{F.}},
\bauthor{\bsnm{Chi}, \binits{N.}}, \betal:
\batitle{Freeform metasurface-assisted waveguide coupler for guided wave polarization manipulation and spin--orbit angular momentum conversion}.
\bjtitle{ACS Photonics}
\bvolume{11}(\bissue{3}),
\bfpage{1051}--\blpage{1059}
(\byear{2024})
\end{barticle}
\endbibitem

\bibitem[\protect\citeauthoryear{Ho et~al.}{2020}]{ho2020denoising}
\begin{barticle}
\bauthor{\bsnm{Ho}, \binits{J.}},
\bauthor{\bsnm{Jain}, \binits{A.}},
\bauthor{\bsnm{Abbeel}, \binits{P.}}:
\batitle{Denoising diffusion probabilistic models}.
\bjtitle{Advances in neural information processing systems}
\bvolume{33},
\bfpage{6840}--\blpage{6851}
(\byear{2020})
\end{barticle}
\endbibitem

\bibitem[\protect\citeauthoryear{Gupta and Brandstetter}{2022}]{gupta2022towards}
\begin{botherref}
\oauthor{\bsnm{Gupta}, \binits{J.K.}},
\oauthor{\bsnm{Brandstetter}, \binits{J.}}:
Towards multi-spatiotemporal-scale generalized pde modeling.
arXiv preprint arXiv:2209.15616
(2022)
\end{botherref}
\endbibitem

\bibitem[\protect\citeauthoryear{Hughes et~al.}{2019}]{hughes2019forward}
\begin{barticle}
\bauthor{\bsnm{Hughes}, \binits{T.W.}},
\bauthor{\bsnm{Williamson}, \binits{I.A.}},
\bauthor{\bsnm{Minkov}, \binits{M.}},
\bauthor{\bsnm{Fan}, \binits{S.}}:
\batitle{Forward-mode differentiation of maxwell’s equations}.
\bjtitle{ACS Photonics}
\bvolume{6}(\bissue{11}),
\bfpage{3010}--\blpage{3016}
(\byear{2019})
\end{barticle}
\endbibitem

\bibitem[\protect\citeauthoryear{Takamoto et~al.}{2023}]{takamoto2023learning}
\begin{bchapter}
\bauthor{\bsnm{Takamoto}, \binits{M.}},
\bauthor{\bsnm{Alesiani}, \binits{F.}},
\bauthor{\bsnm{Niepert}, \binits{M.}}:
\bctitle{Learning neural pde solvers with parameter-guided channel attention}.
In: \bbtitle{International Conference on Machine Learning},
pp. \bfpage{33448}--\blpage{33467}
(\byear{2023}).
\bcomment{PMLR}
\end{bchapter}
\endbibitem

\bibitem[\protect\citeauthoryear{Yu and Capasso}{2014}]{yu2014flat}
\begin{barticle}
\bauthor{\bsnm{Yu}, \binits{N.}},
\bauthor{\bsnm{Capasso}, \binits{F.}}:
\batitle{Flat optics with designer metasurfaces}.
\bjtitle{Nature materials}
\bvolume{13}(\bissue{2}),
\bfpage{139}--\blpage{150}
(\byear{2014})
\end{barticle}
\endbibitem

\bibitem[\protect\citeauthoryear{Li et~al.}{2020}]{li20203d}
\begin{barticle}
\bauthor{\bsnm{Li}, \binits{H.}},
\bauthor{\bsnm{Wang}, \binits{G.-M.}},
\bauthor{\bsnm{Hu}, \binits{G.}},
\bauthor{\bsnm{Cai}, \binits{T.}},
\bauthor{\bsnm{Qiu}, \binits{C.-W.}},
\bauthor{\bsnm{Xu}, \binits{H.-X.}}:
\batitle{3d-printed curved metasurface with multifunctional wavefronts}.
\bjtitle{Advanced Optical Materials}
\bvolume{8}(\bissue{15}),
\bfpage{2000129}
(\byear{2020})
\end{barticle}
\endbibitem

\bibitem[\protect\citeauthoryear{Lee et~al.}{2018}]{lee2018metasurface}
\begin{barticle}
\bauthor{\bsnm{Lee}, \binits{G.-Y.}},
\bauthor{\bsnm{Hong}, \binits{J.-Y.}},
\bauthor{\bsnm{Hwang}, \binits{S.}},
\bauthor{\bsnm{Moon}, \binits{S.}},
\bauthor{\bsnm{Kang}, \binits{H.}},
\bauthor{\bsnm{Jeon}, \binits{S.}},
\bauthor{\bsnm{Kim}, \binits{H.}},
\bauthor{\bsnm{Jeong}, \binits{J.-H.}},
\bauthor{\bsnm{Lee}, \binits{B.}}:
\batitle{Metasurface eyepiece for augmented reality}.
\bjtitle{Nature communications}
\bvolume{9}(\bissue{1}),
\bfpage{4562}
(\byear{2018})
\end{barticle}
\endbibitem

\bibitem[\protect\citeauthoryear{Yang et~al.}{2024}]{yang2024titanium}
\begin{barticle}
\bauthor{\bsnm{Yang}, \binits{J.}},
\bauthor{\bsnm{Van~Gasse}, \binits{K.}},
\bauthor{\bsnm{Lukin}, \binits{D.M.}},
\bauthor{\bsnm{Guidry}, \binits{M.A.}},
\bauthor{\bsnm{Ahn}, \binits{G.H.}},
\bauthor{\bsnm{White}, \binits{A.D.}},
\bauthor{\bsnm{Vu{\v{c}}kovi{\'c}}, \binits{J.}}:
\batitle{Titanium: sapphire-on-insulator integrated lasers and amplifiers}.
\bjtitle{Nature}
\bvolume{630}(\bissue{8018}),
\bfpage{853}--\blpage{859}
(\byear{2024})
\end{barticle}
\endbibitem

\bibitem[\protect\citeauthoryear{Jang et~al.}{2025}]{jang2025inverse}
\begin{botherref}
\oauthor{\bsnm{Jang}, \binits{E.}},
\oauthor{\bsnm{Cho}, \binits{J.}},
\oauthor{\bsnm{Kang}, \binits{C.}},
\oauthor{\bsnm{Chung}, \binits{H.}}:
Inverse design of ultrathin metamaterial absorber.
arXiv preprint arXiv:2504.14901
(2025)
\end{botherref}
\endbibitem

\bibitem[\protect\citeauthoryear{Roques-Carmes et~al.}{2022}]{roques2022toward}
\begin{barticle}
\bauthor{\bsnm{Roques-Carmes}, \binits{C.}},
\bauthor{\bsnm{Lin}, \binits{Z.}},
\bauthor{\bsnm{Christiansen}, \binits{R.E.}},
\bauthor{\bsnm{Salamin}, \binits{Y.}},
\bauthor{\bsnm{Kooi}, \binits{S.E.}},
\bauthor{\bsnm{Joannopoulos}, \binits{J.D.}},
\bauthor{\bsnm{Johnson}, \binits{S.G.}},
\bauthor{\bsnm{Soljacic}, \binits{M.}}:
\batitle{Toward 3d-printed inverse-designed metaoptics}.
\bjtitle{Acs Photonics}
\bvolume{9}(\bissue{1}),
\bfpage{43}--\blpage{51}
(\byear{2022})
\end{barticle}
\endbibitem

\bibitem[\protect\citeauthoryear{Fan et~al.}{2024}]{fan20243d}
\begin{barticle}
\bauthor{\bsnm{Fan}, \binits{D.}},
\bauthor{\bsnm{Smith}, \binits{C.S.}},
\bauthor{\bsnm{Unnithan}, \binits{R.R.}},
\bauthor{\bsnm{Kim}, \binits{S.}}:
\batitle{3d printed diffractive optical elements for rapid prototyping}.
\bjtitle{Micro and Nano Engineering}
\bvolume{24},
\bfpage{100270}
(\byear{2024})
\end{barticle}
\endbibitem

\bibitem[\protect\citeauthoryear{Hendrycks and Gimpel}{2016}]{hendrycks2016gaussian}
\begin{botherref}
\oauthor{\bsnm{Hendrycks}, \binits{D.}},
\oauthor{\bsnm{Gimpel}, \binits{K.}}:
Gaussian error linear units (gelus).
arXiv preprint arXiv:1606.08415
(2016)
\end{botherref}
\endbibitem

\bibitem[\protect\citeauthoryear{Loshchilov and Hutter}{2017}]{loshchilov2017decoupled}
\begin{botherref}
\oauthor{\bsnm{Loshchilov}, \binits{I.}},
\oauthor{\bsnm{Hutter}, \binits{F.}}:
Decoupled weight decay regularization.
arXiv preprint arXiv:1711.05101
(2017)
\end{botherref}
\endbibitem

\bibitem[\protect\citeauthoryear{Loshchilov and Hutter}{2016}]{loshchilov2016sgdr}
\begin{botherref}
\oauthor{\bsnm{Loshchilov}, \binits{I.}},
\oauthor{\bsnm{Hutter}, \binits{F.}}:
Sgdr: Stochastic gradient descent with warm restarts.
arXiv preprint arXiv:1608.03983
(2016)
\end{botherref}
\endbibitem

\bibitem[\protect\citeauthoryear{Paszke}{2019}]{paszke2019pytorch}
\begin{botherref}
\oauthor{\bsnm{Paszke}, \binits{A.}}:
Pytorch: An imperative style, high-performance deep learning library.
arXiv preprint arXiv:1912.01703
(2019)
\end{botherref}
\endbibitem

\end{thebibliography}

\section{Acknowledgements}
This research was funded by the National Research Foundation of Korea (NRF) grant funded by the Korean government (MSIT) under the grant numbers RS-2024-00338048 and RS-2024-00414119. It was also supported by the Global Research Support Program in the Digital Field (RS-2024-00412644) under the supervision of the Institute of Information and Communications Technology Planning \& Evaluation (IITP), and by the Artificial Intelligence Graduate School Program (RS-2020-II201373, Hanyang University), also supervised by the IITP. Additionally, this research was supported by the Artificial Intelligence Semiconductor Support Program (IITP-(2025)-RS-2023-00253914), funded by the IITP, and by the Korean government (MSIT) under the grant numbers RS-2023-00261368, RS-2025-02218723, and RS-2025-02283217. This work also received support from the Culture, Sports and Tourism R\&D Program through a grant from the Korea Creative Content Agency (KOCCA), funded by the Ministry of Culture, Sports and Tourism (RS-2024-00332210). The APC was funded by the same sources.

\section{Competing interests}
The authors declare no competing interests.

\section{Author contributions}
This project was initiated by J.S. and C.K. The entire simulations were designed by C.K., with assistance from H.C. C.K. implemented the simulation codes and conducted data collection, supported by J.S. Model training was performed by J.S. Experiment design was carried out by J.S. and C.K., with assistance from D.S. All experiments were conducted by J.S., with support from C.K. The manuscript was prepared by J.S. and C.K., with assistance from H.C.

The work was initially conceived by J.S and C.K, and designed with contributions from A.T. A.F. developed the machine learning code, performed the model training, and analyzed the model performance in diverse applications together with L.M.S. A.T. supervised and revised all stages of the work. All authors discussed the results and contributed to the final manuscript.

\end{document}